\documentclass[journal]{IEEEtran}

\usepackage{graphicx}
\usepackage{multirow}
\usepackage{makecell}
\usepackage{booktabs}
\usepackage{caption}
\usepackage{wrapfig}
\usepackage{amsmath}
\usepackage{color}
\usepackage[rightcaption]{sidecap}
\usepackage{bm}

\begin{document}

\title{Exemplar-Based Image Colorization with A Learning Framework}
\author{Zhenfeng Xue \textsuperscript{$\dagger,\ddagger$},
		Jiandang Yang \textsuperscript{$\dagger$},
		Jie Ren \textsuperscript{$\dagger,\ddagger$},
		Yong Liu \textsuperscript{$\dagger,\ddagger$}
\thanks{$^{\dagger}$Z. Xue, J. Yang, J. Ren and Y. Liu are with the School of Control Science and Engineering, Zhejiang University, No.38 Zheda Road, Hangzhou, China. Yong Liu is the corresponding author. (E-mail: zfxue0903@zju.edu.cn, yangjd@zju.edu.cn, renjie@bdcatek.com, yongliu@iipc.zju.edu.cn)}
\thanks{$^{\ddagger}$Z. Xue, J. Ren and Y. Liu are also with the Research Center for Intelligent Perception and Control, Huzhou Institute of Zhejiang University, No.819 Xisaishan Road, Huzhou, China.}
}


\maketitle

\begin{abstract}

Image learning and colorization are hot spots in multimedia domain. Inspired by the learning capability of humans, in this paper, we propose an automatic colorization method with a learning framework. This method can be viewed as a hybrid of exemplar-based and learning-based method, and it decouples the colorization process and learning process so as to generate various color styles for the same gray image. The matching process in the exemplar-based colorization method can be regarded as a parameterized function, and we employ a large amount of color images as the training samples to fit the parameters. During the training process, the color images are the ground truths, and we learn the optimal parameters for the matching process by minimizing the errors in terms of the parameters for the matching function. To deal with images with various compositions, a global feature is introduced, which can be used to classify the images with respect to their compositions, and then learn the optimal matching parameters for each image category individually. What's more, a spatial consistency based post-processing is design to smooth the extracted color information from the reference image to remove matching errors. Extensive experiments are conducted to verify the effectiveness of the method, and it achieves comparable performance against the state-of-the-art colorization algorithms.

\end{abstract}

\begin{IEEEkeywords}
image colorization, feature matching, parameter optimization, learning-based method
\end{IEEEkeywords}

%
\IEEEpeerreviewmaketitle

\section{Introduction}

Image based learning and recognition, such as image segmentation~\cite{DBLP:journals/tsmc/LiGKH17,DBLP:journals/tsmc/PengCYT17,yang2015graph,peng2016high,sima2017bottom}, image retrieval~\cite{DBLP:journals/tsmc/PassalisT17,zhang2012automatic,liu2016unsupervised,zhang2017landmark} and image colorization~\cite{zong2015fast,li2017example} are important hot spots in cybernation field. Among them, image colorization aims to recover a color image from a grayscale image, which is widely used in the image editing cases, such as generating pictures with different color styles, restoring the colors of the old pictures or grayscale movies~\cite{zong2015fast,li2017example,Zhang:2017:RUI:3072959.3073703,he2018deep}. Although the color information of those gray images are incomplete, the humans are still able to deduce the missing information based on their prior knowledge and some colourful scene examples with similar image contents. The starting point of this paper is inspired by the learning capability of humans.

As there are two channels of the color information are missing, the colorization process is a typical ill-posed problem, and thus additional information is desired to solve this problem. There are three main ways to provide those additional information, \textit{i.e.}, interaction-based, exemplar-based and learning-based methods. The interaction-based methods~\cite{Levin:2004,Yatziv:2004,Huang:2005,Luan:2007,zong2015fast} require to scribe a few colors on the target image, and then using the optimization method to smoothly spread the manual annotation color values to the entire target image to complete the colorization. The disadvantage lies in that it costs too much manual work. In addition, the colorization results are highly relying on the color values scribed by the manual work, which requires highly professional skills.

The exemplar-based methods~\cite{Welsh:2002,Irony:2005,Charpiat:2008,Liu:2008,Gupta:2012,pierre2015luminance,li2017example} employ a reference image similar to the grayscale image to provide additional information. Then they can transfer the color information from the reference image to the target image by a matching function. In most cases, these approaches can reduce the manual efforts, but the performance is closely related to the composition of the images that need to be colorized. The colorization performance may vary with the change of image composition.

The learning-based methods~\cite{deshpande2015learning,varga2016fully,larsson2016learning,
zhang2016colorful,iizuka2016let,cheng2017colorization,zhang2017real,
xiao2019interactive} use large-scale training samples to encode the relations between scenes and objects of the ground truth color information, and then apply the encoded information to the grayscale image. Although these methods do not need any manual scribble or reference image, they are hard in training and time-costly. More importantly, the colorization results highly depend on the training data and are unstable. Once the training data could not successfully contain those colorful images, the colorization results will become unreasonable. Furthermore, there exists a tight coupling relationship between the colorization style and learning process. That is to say, it can only generate one unique colorization result due to their tightly coupled training models, and they are hard to transfer image to various colorization styles.

Concerning the above mentioned weaknesses of current colorization methods, this paper proposes an automatic exemplar-based colorization method with a learning framework~\cite{han2016two}. It can be viewed as a hybrid method of exemplar-based and learning-based method. The basic idea of our approach is to regard the matching process in the exemplar-based method as a parameterized function, and we employ a large amount of color images as the training samples to fit the parameters. We can learn the optimal parameters for the matching process in colorization by minimizing the errors between colorizaiton results and the ground truth image in terms of the parameters. To deal with images with various compositions, we introduce a global feature module, which can be used to classify those large amount images with respect to their compositions, and then learn the optimal matching parameters for each image category individually. To remove the possible matching errors introduced by wrong information, we further propose a spatial consistency-based post-porocessing module to smooth the extracted color information from the reference image.

In summary, the contributions are listed as follows.

\begin{itemize}
	\item A scalable exemplar-based learning framework is proposed for image colorization, which decouples the colorization styles and learning process by integrating both exemplar-based and learning-based method.
	
	\item A global feature module is introduced to cluster the images into multiple categories so that different matching parameters can be learned for each category to solve the difficulty caused by image composition variation.
	
	\item We perform extensive experiments to verify the effectiveness of the method, and it achieves comparable performance against existing state-of-the-art methods.

\end{itemize}

\section{Related work}

The colorization methods can be categorized into three main kinds, \textit{i.e.}, manual interaction-based method, automatic exemplar-based method and learning-based method.
The \textbf{first} kind of methods need some manual work when colorizing the images. For example, Levin \textit{et al.}~\cite{Levin:2004} propose a simple yet effective algorithm based on the relationship between luminance and chrominance. The method requires users to provide the color values on various areas of the image,  and then the color can be spread to the entire image automatically via minimizing the mean square error. Yatziv \textit{et al.}~\cite{Yatziv:2004} employ multiple manual color scribbles to colorize the images, where the weights of individual color are decided by the distance function between pixels. These manual interaction methods can perform well in condition that those manually annotated colors are input correctly. 

To solve the difficulty of obtaining color values directly from the manual scribbles, the \textbf{second} kind of methods try to make the colorization process easier by getting the color values from the user-supplied images automatically. For example, Welsh \textit{et al.}~\cite{Welsh:2002} propose to extract image blocks from the target grayscale image, and then match with the reference image based on the statistic information of pixels. As their approach ignores the spatial information of pixels, the results in local details usually cannot behave very well.

To deal with the problem of spatial coherence, Irony \textit{et al.}~\cite{Irony:2005} propose a new method to use the segmentation map of reference image as an additional input. They utilize the colors and textures of the segmented regions from the reference image to transfer the color values to target grayscale image. Then the colors are spreading to the entire image by the interpolation algorithm. As their method requests the segmentation mask as the input, the colorization results highly depend on the quality of the segmentation. Another kind of exemplar-based colorization~\cite{Gupta:2012,Bugeau:2014} methods try to combine multiple patch based features and also consider the spatial consistency during the colorization process.

The work completed by~\cite{Gupta:2012} is the most relative work compared with ours. The approach extracts super-pixels from the target image and the reference image. Color values can be obtained via the nearest neighbour matching, and then adjust the color values by image space voting. This method never considers the influence of the polytropism of compositions on the scenes and objects in images and employed an unified empirical matching parameters for all the images in color transferring. As a result, their method can only work well in some images with specific compositions.

The \textbf{third} kind of methods~\cite{cheng2015deep}~\cite{zhang2016colorful}  are recently emerged works. For example, Cheng \textit{et al.}~\cite{cheng2015deep} propose a DNN (deep neural network) to train on a large set of reference images and use the pre-trained neural network to colorize the grayscale image.  Zhang \textit{et al.}~\cite{zhang2016colorful} propose a CNN architecture to train on 1.3 million images from ImageNet~\cite{russakovsky2015imagenet}. They use the grayscale images which contain only $l$ channel in $l\alpha\beta$ color space as input, and then quantize the $\alpha\beta$ channels into 313 bins to convert the colorization process into a classification problem. This category of methods are supposed to be trained on a huge image database, and thus they highly depend on the training data set. Once there are no similar images to the target images, the colorization results may become unreasonable. It also cannot output multiple colorization results compared with the exemplar-based methods, as the DNN can outputs only one unique colorization result.
Other similiar works such as~\cite{he2018deep,xiaoexample} combine exemplar-based method and end-to-end learning-based method to encode the grayscale image and reference image on large-scale dataset. For example, in~\cite{he2018deep},  they propose to use two sub-networks, \textit{i.e.}, the similarity sub-net and colorization sub-net, which are trained on 0.7 million image pairs sampled from ImageNet~\cite{russakovsky2015imagenet}.


As we know, the quality of exemplar-based methods highly rely on the structural compositions of the images. That's the reason why the previous exemplar-based methods perform not so good in the images with various structures. Although learning-based methods are able to colorize the images without providing corresponding reference images, the reference images are already potentially encoded into their learned models in fact. That's the reason why the learning-based methods highly depend on their training samples, and there should be highly similar reference color images in the training sets for the target grayscale images. As a result, we propose a hybrid of exemplar-based and learning-based approach to overcome the disadvantages of each other.

\section{Proposed method}

\begin{figure*}[htp]
	\centering
	\includegraphics[scale=0.56]{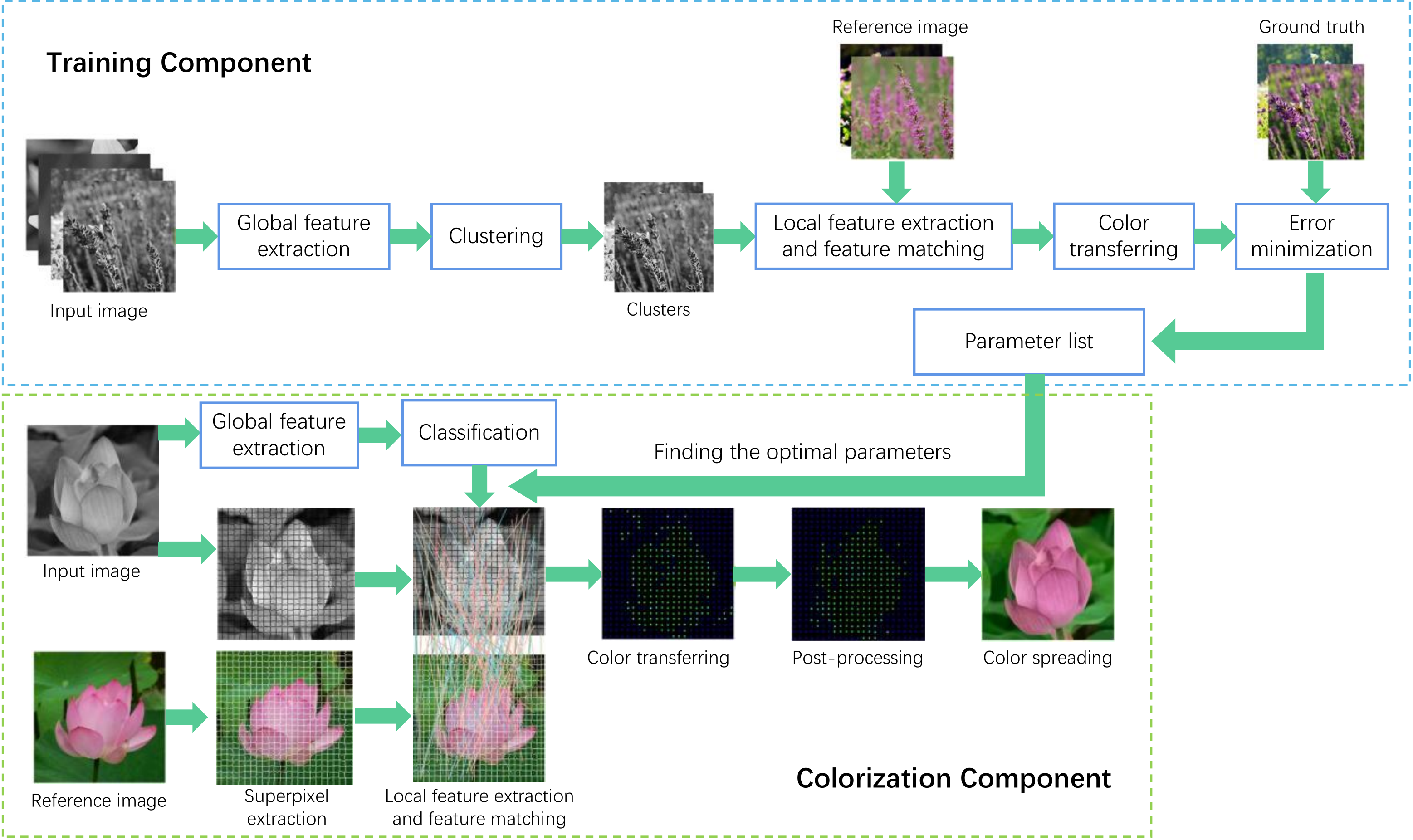}
	\caption{An overview of the proposed framework. In the training component, we group the training samples into several clusters and learn the parameters correspondingly. In the colorization component, we assign each input image with its corresponding parameters determined by global feature, and match the superpixels between gray and reference images by local features.}
	\label{fig:1_overview}
\end{figure*}

In this paper, we propose an exemplar-based image colorization method with a learning framework. The method regards the matching process in exemplar-based method as a parameterized function and we employ the learning-based method to learn the optimal parameters for the matching process. Specifically, a global feature extraction module is introduced to represent global structural compositions of images and then the images are categorized into specific groups. Then, different parameters can be learnt according to the categorical information, and the method is able to adapt to images with different structural compositions. The overall framework is illustrated in Fig.~\ref{fig:1_overview}.

In the feature extraction component, two feature extraction modules are introduced, \textit{i.e.}, the global and local feature modules. The former represents the global structural composition of input image and is used to cluster the samples. Here, the input gray image is assumed to share the same parameters with the training cluster, where the clustering center is closest to the input image in terms of their global features. The local features are used to describe the local patch similarity between the grayscale and reference images.

\begin{figure}[htp]
	\centering
	\includegraphics[scale=0.4]{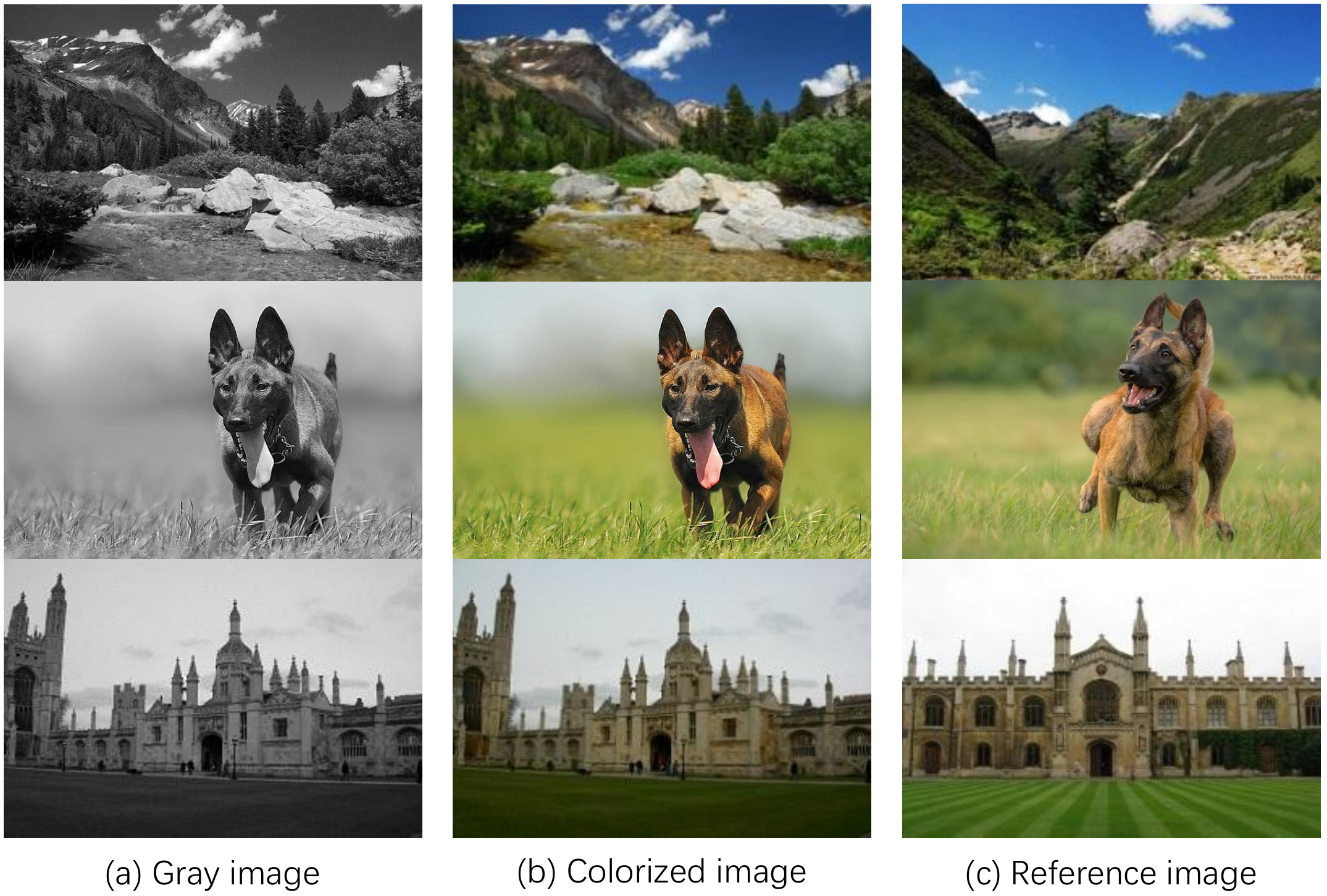}
	\caption{Example training pairs in our method, in which thee images in a row consist of a training pair.}
	\label{fig:2_reference}
\end{figure}

The overall framework is split into two main stages, \textit{i.e.}, the training stage and the colorization stage. In the training component, we learn the matching parameters for each cluster, which is to find the most optimal matching parameters from the reference image. The training samples are constructed by the color images contained in each cluster. As shown in Fig.~\ref{fig:2_reference}, each training sample consists of three images, \textit{i.e.}, the gray image, colorized image and reference image. We use parameterized local features to construct patch-based local descriptors for the gray image and reference image. To obtain the color transferring image, we transfer the colors from the reference image to gray image by their descriptor similarity. The optimal matching parameters for each cluster can be calculated by minimizing the color difference between ground truth and generated images.

In the colorization component, we firstly choose the proper local feature parameters for the input gray image based on its global feature category. Then the parameterized local features are used to obtain the color transferring image from its reference image. We also propose a post-processing operation for the generated color image to remove isolations or abnormal color transfer. The final colorization results can be obtained by the color interpolation algorithm~\cite{Levin:2004}.

\subsection{Feature Extraction}

\subsubsection{Global Feature Extraction}

In order to represent the global structural composition of images, we propose to use GLCM (Grey Level Co-occurrence Matrix)~\cite{Haralick:1973} to extract global features. The reasons are multifacted. Firstly, GLCM has the advantage of reflecting the direction, amplitude, and neighborhood of the grayscale image. Meanwhile, GLCM can reflect the distribution characteristics of pixels with the same gray levels. Secondly, GLCM is a tabulation that reflects the frequency of different combinations of pixel brightness values in an image. Thus, the global features constructed on GLCM are able to describe the image texture features for the input images with different complexity and composition.

\begin{figure}[htp]
	\centering
	\includegraphics[scale=0.5]{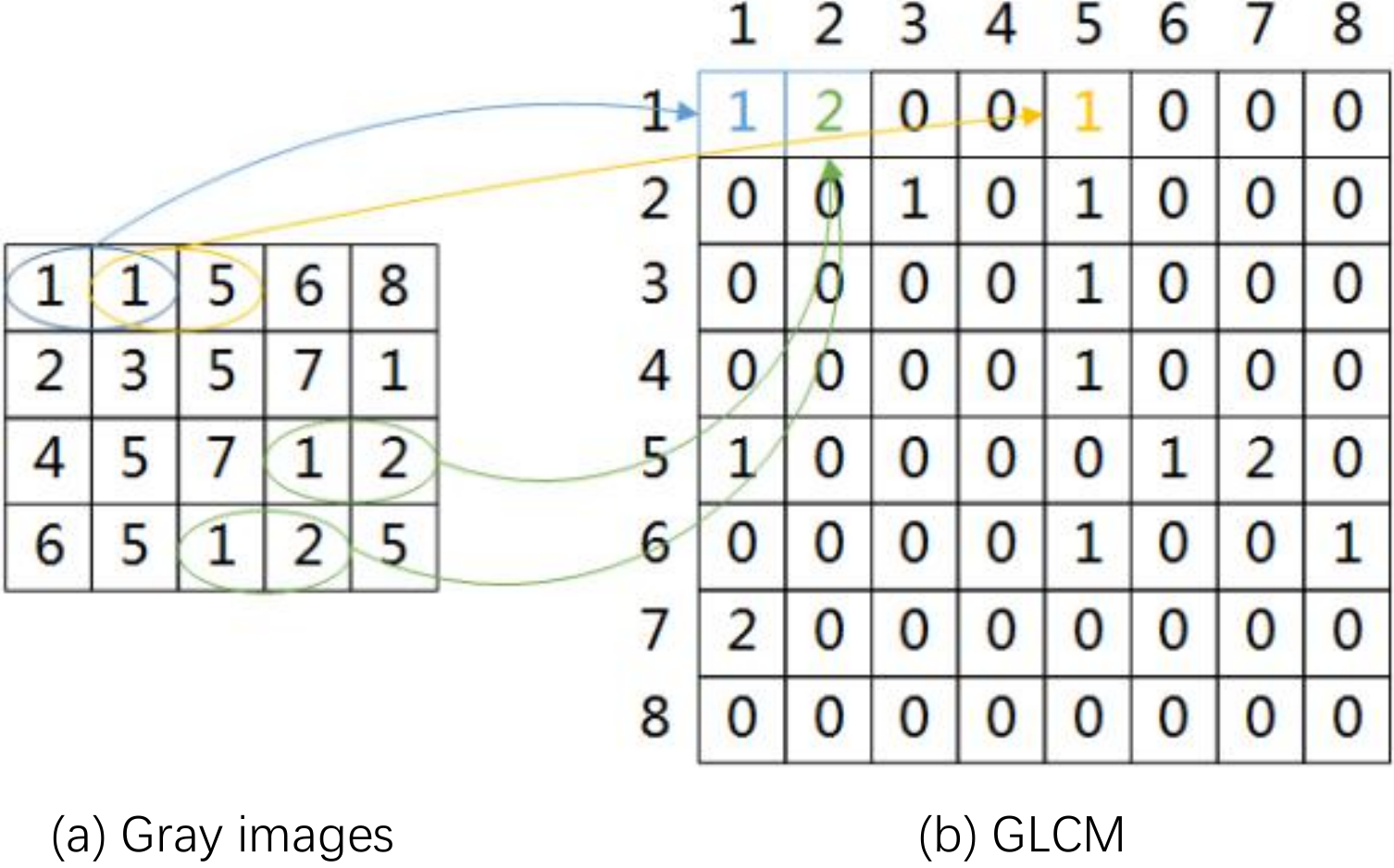}
	\caption{An illustration of GLCM construction process. Here, only the horizontal GLCM is calculated, and each element shows the co-occurrence of a gray level pair.}
	\label{fig:3_glcm}
\end{figure}

A simple calculation process of GLCM is shown in Fig. \ref{fig:3_glcm}. The left shows a grayscale quantized image whose gray level is 8. The right shows the corresponding GLCM obtained from the left image. The matrix is a 8$\times$8 matrix with each element equaling the frequency of corresponding gray level pair in the image. More details can be found in~\cite{Haralick:1973}.

In order to suppress the directional component, we dispose the results by calculating the GLCM in four different directions ($0^\circ$, $45^\circ$, $90^\circ$, $135^\circ$). Then we construct four textural features which can be extracted from each of the GLCM.

\begin{itemize}
\item Angular Second Moment
\begin{equation}
\textrm{ASM}=\sum_{i}\sum_{j}P\left(i,j\right)^{2},
\end{equation}
\item Contrast
\begin{equation}
\textrm{CON}=\sum_{i}\sum_{j}\left(i-j\right)^{2}P\left(i,j\right),
\end{equation}
\item Correlation
\begin{equation}
\resizebox{.78\hsize}{!}{$\textrm{CORRLN}=\left[\sum_{i}\sum_{j}\left(\left(ij\right)P\left(i,j\right)\right)-\mu_{x}\mu_{y}\right]/\sigma_{x}\sigma_{y}$},
\end{equation}
\item Entropy
\begin{equation}
\textrm{ENT}=-\sum_{i}\sum_{j}P\left(i,j\right)\log P\left(i,j\right),
\end{equation}
\end{itemize}
where $P\left(i,j\right)$ denotes the GLCM, $\mu_x$, $\mu_y$, $\sigma_x$ and $\sigma_y$ are the means and standard deviations of the marginal distributions associated with $P(i, j) / R$ and $R$ is a normalization constant. The global features are averaged over different GLCMs.

Before learning the matching parameters for the local features, we need to extract the global feature for each input image, which is a four-dimensional vector corresponding to the four textural feature mentioned above. We then use the AP (Affinity Propagation) algorithm~\cite{Xiao:2007} to cluster the target images based on the global features. The training process can be parallelly executed on each category after clustering that is illustrated in Section~\ref{section:training}.

\subsubsection{Local Feature Extraction}
The local feature extraction is established on the level of super-pixel. The reason is that super-pixel can strengthen the robustness against the pixel noise in the colorization process, and helps to reduce the searching space. Compared to individual pixel, super-pixel has prominent advantage regarding the image spatial relatedness. Here, we extract the super-pixels of the target image and the reference image based on geometric flow~\cite{Levinshtein:2009}. The super-pixels extracted by the algorithm possess equalization on both the size and shape. This ensures the connectedness and the compactness, and retains the original image border. As common practise, the number of pixels in each super-pixel is around 100 in all of the experiments.

After obtaining the super-pixels of target image and reference image, four local properties are used, including intensity, standard deviation, Gabor and SURF features.

\begin{itemize}
\item Intensity. The $l$ channel of every pixel in $l\alpha\beta$ color space is used to describe the pixel intensity feature. The super-pixel intensity feature is the average value of all the pixels within the super-pixel.The computational formula is:
\begin{equation}
F_1=\frac{1}{n}\sum_{i\leq n}I_i.
\end{equation}

\item Standard deviation. In order to obtain the neighborhood spacial information of pixels, we apply a 5$\times$5 pixel template to calculate the standard deviation for each pixel within the super-pixel. The super-pixel standard deviation is defined as the mean value of all the pixels' standard deviations within the super-pixel.

\item Gabor feature~\cite{Manjunath:1996}. Gabor feature is effective in describing the image texture information. We use Gabor filter to extract a 40-dimensional feature include 8 directions and 5 exponential scales. Then we compute the average value of pixels within the super-pixel as the Gabor feature.

\item SURF descriptor~\cite{Bay:2008}. We extract the 128-dimensional feature for each pixel, and then compute the average value for all the pixels in the super-pixel.
\end{itemize}

\subsection{Training for Matching Parameters}~\label{section:training}
The training component is to learn the weight parameters for the local features from a large number of color images. Transferring the color images into grayscale images can get a large amount of training samples. As unified weight parameters may not be fulfilled with all kind of images, we then cluster the training images into several categories and learn different optimal weight parameters for each category. The main steps include \emph{feature matching}, \emph{error calculation} and \emph{parameter optimization}.

\subsubsection{Feature Matching}

Feature matching is to find a super-pixel in the reference image whose local feature vector is closest to the super-pixel in the target image. The aim of this process is to obtain all the corresponding relation between the reference image and the target image in the super-pixel level. Then we can transfer the central pixel color of the super-pixel from the reference image to the target image based on the corresponding relation of those super-pixels, and obtain the color transferring image~\footnote{In our approach, only the central pixel's color of the super-pixel is transferred, other colors of the target image will be recovered by the color spreading in section~\ref{colorspreading}}. For the reason that the three color channels are basically perpendicular and with small correlation in $l\alpha\beta$ color space, we only need transfer the $\alpha$ and $\beta$ channels from the reference image as the color values.

We use the following parameterized similarity function to denote the distance between two super-pixels:
\begin{equation}
G\left(W,t_{i},r_{j}\right)=\sum_{k=1}^{K}w_{k}\left\|F_{k}^{t_{i}}-F_{k}^{r_{j}}\right\|\left(t_{i}\in{T},r_{j}\in{R}\right),
\label{equation:similarity}
\end{equation}
where $T$ represents the target image super-pixel set, $R$ represents the reference image super-pixel set. $t_{i}$ and $r_{j}$ are elements from $T$ and $R$. $W=\left(w_{1},w_{2},\cdots,w_{k}\right)$ represents the weight parameters of local features $\left(F_{1},F_{2},\cdots,F_{k}\right)$, which is varied based on the target image's global feature. We use the local features defined in Section 4.2,  and  then $K$=4. $\left\|\cdot\right\|$ denotes L2-norm.

The process of super-pixel matching can be formally described as, $\forall t_{i}\in T$, find $r_{j}$, $s.t.$, $\forall r_{k}\in R$, satisfying:
\begin{equation}
G\left(W,t_{i},r_{j}\right)\leq G\left(W,t_{i},r_{k}\right).
\label{equation3}
\end{equation}

We can get $t_{i}$'s corresponding super-pixel $r_{j}$ through formula~(\ref{equation3}). Then the color of $r_{j}$'s central pixel will be transferred to $t_{i}$'s central pixel. For convenience, we use $C\left(\cdot\right)$ to denote the transferred color of super-pixel. Where $C\left(\cdot\right)$ is a 2-dimensional vector that represents the $\alpha\beta$ values in $l\alpha\beta$ color space. Then we rewrite the color transfer process as $C\left(t_{i}\right)\leftarrow C\left(r_{j}\right)$.  After matching features and transferring colors for every super-pixels in the target image, we get the color transferring image.

\subsubsection{Error Calculation and Parameters Optimization}

The formula~(\ref{equation:similarity}) shows the similarities of the super-pixels are relying on the local feature and the weight parameters, which will directly lead to varied colorization results. On the other side, it is also hard to design an unified similarity function to represent images with different global structural compositions. Thus we regard $W$ as the weight parameters to control the similarity function and to construct different similarity functions for different categories of images. For each category clustering by the global feature, we need to calculate its corresponding optimal weight parameters $W$.

The error function of a sample set is defined as:
\begin{equation}
Error\left(T\right)=\sum_{t_{i}\in T}\left\|C\left(\check{t}_{i}\right)-C\left(t_{i}\right)\right\|,
\end{equation}
where $C\left(\check{t}_{i}\right)$ represents the color value of the ground true image.
Based on the formula~(\ref{equation3}), We use $C\left(t_{i},R,W\right)$ to present the color value of the super-pixel found in $R$ which is most similar with $t_{i}$ by giving a certain $W$.
Then the error function can be adapted into:
\begin{equation}
Error\left(T,R,W\right)=\sum_{t_{i}\in T}\left\|C\left(t_{i},R,W\right)-C\left(\check{t}_{i}\right)\right\|.
\end{equation}
The goal of parameter optimization is to get an optimal weight combination of the local features for each category of images. That is equal to find an optimum set of attribute weights $\overrightarrow{W}$ to minimize the error function. Therefore, optimal weights $\overrightarrow{W}$ is to minimize the sum of error of a category:
\begin{equation}
\overrightarrow{W}=\arg\min_{W}\sum_{n}Error\left(T^{n},R^{n},W\right)
\end{equation}
{where $n$ is the sample number contained in the corresponding category, here each category is clustered from the trainning sample set by the global feature GLCM.}

The above equation is a process optimization problem which can be solved by Levenberg-Marquardt~\cite{Manolis:2005}. And the initial value of $W$ is given randomly.

\subsection{Colorization from Reference Image}

The colorization component is to colorize the target image according to the reference image. Firstly, we choose the proper weight parameters for the input target image. That is to find the category whose cluster center is closest to the input target image on the global feature vector, and then to assign the pre-trained weight parameters of that category to the input target image. Then, we use those selected parameters to find the most matched super-pixel from the reference image and then obtain the input target image's transferring image.

Before applying the color spreading method to colorize the remainder pixels of the input target image, we introduce a post-processing with spatial consistency to improve the result of the color transferring image obtained by matching from reference images.

\subsubsection{Post-Processing with Spatial Consistency}

As the target image $T$ is divided into a non-overlapping and full-covered super-pixel set in our approach, and then each super-pixel $t_i$ can be regarded as a node in a graph.
And the neighborhood relations among those super-pixels consist of the edges of the graph. We then introduce the CRF (Conditional Random Fields)~\cite{Lafferty01conditionalrandom}, which is a good structure to describe probabilities in a graph, to further optimize the color transferring image and remove those possible wrong matchings in terms of the spatial consistency. The CRF includes an unary potentials $\psi_i$ and the potentials between pairs of nodes $\phi_{ij}$.

We use $\hat{C}(t_i)$ to denote the values of $\alpha, \beta$ transferred from the reference image. Then post-processing is to find an optimal color set $C$ that can maximize the following probability:
\begin{equation}
P(C|\hat{C})=\frac{1}{Z}\prod_{i}{\psi_i}{(\hat{C}(t_i)|C(t_i))}\prod_{\forall i,j, t_i}\phi_{ij}{(C(t_i),C(t_j))}
\end{equation}
where $Z$ is a normalization factor to ensure that the probabilities are in the interval [0,1]. $N^*(t_j)$ is a redefined neighborhood set of $t_j$ according to the intensity and it is standard deviation. We define a new neighborhood relation between two super-pixels: if $t_i$ and $t_j$ are adjacent in the image graph and satisfy $\left| {{l_i} - {l_j}} \right| \le {\delta _1}\& \& \left| {St{d_i} - St{d_j}} \right| \le {\delta _2}$,  and then $t_i$ and $t_j$ are neighbors. As for super-pixel $t_j$, we calculate all it is neighbors and add them to $N^*(t_j)$. Then for all the elements in $N^*(t_j)$, we calculate their neighbors and add them to $N^*(t_j)$ as well. We will repeat this operation until the number of elements in $N^*(t_j)$ up to a constant threshold, {{or the number of repetitions reaches an upper bound}}. Finally, we get $t_j^,$s redefined neighborhood set $N^*(t_j)$.

Here $l_i$ and $l_j$ denote the average intensities of all the pixels within super-pixels $t_i$ and $t_j$ respectively. $Std_i$ denote the standard deviation of all the pixels' intensities in $t_i$. $\delta_1$ and $\delta_2$ are constant thresholds to decide whether the colors of those two adjacent node $t_i$ and $t_j$ should be regard related based on their information on $l$ channel and their standard deviation. In our experiment, we set $\delta_1=0.04$, $\delta_2=0.015$.

The two potentials $\psi_i$ and $\phi_{ij}$ are defined as:
\begin{equation}
{\psi_i}{(\hat{C}(t_i)|C(t_i))}=\exp\left( {-\gamma ||\hat{C}(t_i)-C(t_i)||}\right)
\end{equation}
\begin{equation}
\begin{split}
&{\phi_{ij}}{(C(t_i),C(t_j))}= \\
&\begin{cases}
  \exp\left({-\eta \|C(t_i)-{\sum\limits_{{t_k} \in {N^*}\left( {{t_j}} \right)} {C\left( {{t_k}} \right)} /n}\|}\right)&,\text{ if } {t_i} \in {N^*}\left( {{t_j}} \right) \\
  1&,\text{ if } {t_i} \notin {N^*}\left( {{t_j}} \right)
\end{cases}
\end{split}
\end{equation}
where $\gamma$ and $\eta$ are two nonnegative constants for the balance of these two potentials. $n$ is the number of elements in ${N^*}\left( {{t_j}} \right)$.

\begin{figure}[htp]
\centering
\includegraphics[scale=0.46]{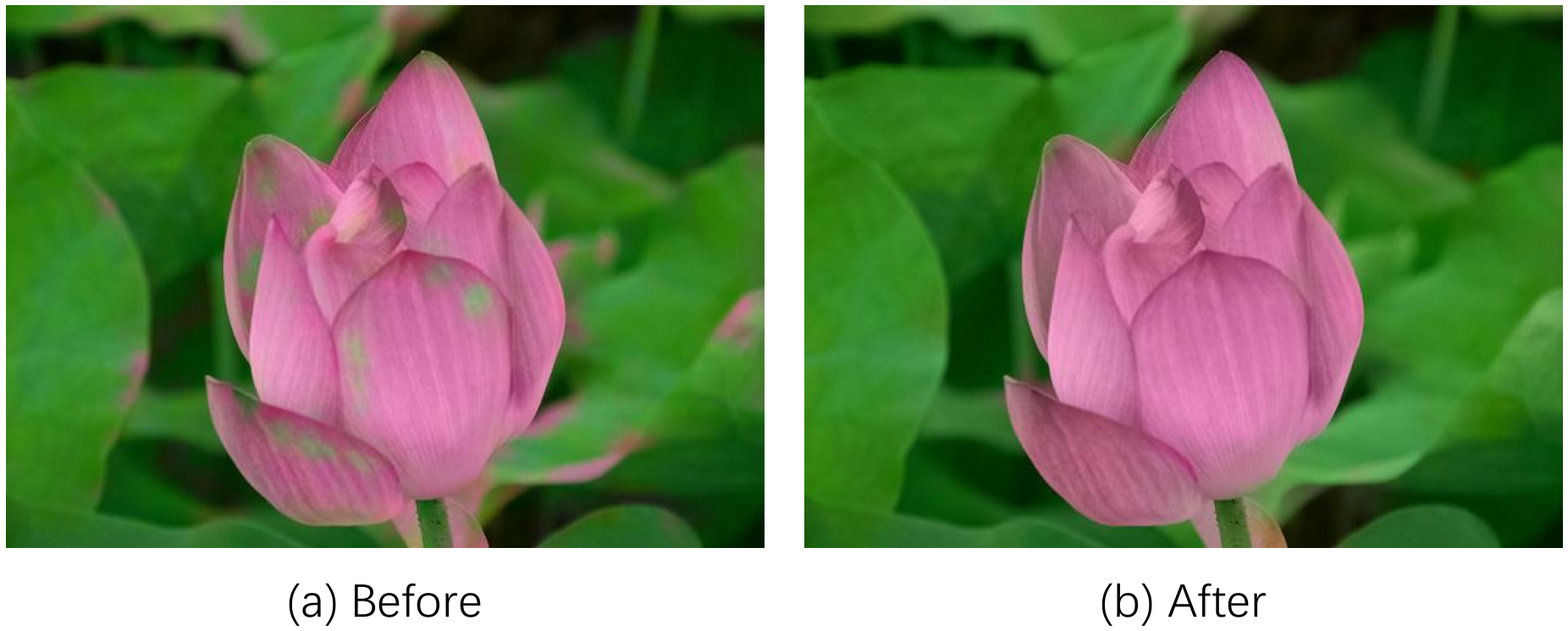}
\caption{(a) Original colorized image. (b) Image after color correction. Comparing (a) and (b), we conclude that the image quality has great improvements through post-processing operation, especially in eliminating noise points.}
\label{bacc}
\end{figure}

A sample result after our CRF based post-processing is given in Fig.~\ref{bacc}. And the results show those wrong matchings are corrected by our post-processing.

\subsubsection{Color Spreading}~\label{colorspreading}

After post-processing, we get a more reasonable color transferring image. In the $l\alpha\beta$ color space, the first channel of this image is the luminance values of each pixel. While the second and the third channels have got chromatic values only in the centers of the super-pixels. We can use the color interpolation algorithms~\cite{Levin:2004} to spread the color to each pixel in the whole image. This algorithm is based on a simple assumption: if two adjacent pixels have similar intensities, and then they have similar color values. The chromatic values in the centers of the super-pixels which are transformed from the reference image are treated as the known values. We optimize the $\alpha\beta$ channels of the other pixels within the image. According to the aforementioned assumption, the color distribution turns into the process of minimizing the difference between pixel's color values and the weighted average color values of the neighboring pixels:
\begin{equation}
J\left(C\right)=\sum_{i=1}^{n}\left(C\left(i\right)-\sum_{j\in N\left(i\right)}w_{ij}C\left(j\right)\right)^{2},
\end{equation}
where $C = \left[ \begin{array}{l}
\alpha\\
\beta
\end{array} \right]
$ denotes the $\alpha\beta$ channel in $l\alpha\beta$ color space. $C\left(i\right)$ and $C\left(j\right)$ are the color of pixels $i$ and $j$. $n$ is the total number of pixels within the image. $N\left(i\right)$ is the neighborhood pixel set of $i$. We define two pixels ${p_1}\left( {{x_1},{y_1}} \right)$ and ${p_2}\left( {{x_2},{y_2}} \right)$ are neighbors if their coordinates satisfy the constraint $\left| {{x_1} - {x_2}} \right| \le 1\& \& \left| {{y_1} - {y_2}} \right| \le 1$. $w_{ij}$ is the weight between $i$ and $j$ according to the intensities:
\begin{equation}
w_{ij}\propto\exp\left(-\left(l\left(j\right)-l\left(i\right)\right)^{2}/2\sigma_{i}^{2}\right)
\end{equation}
where $l\left(i\right)$ and $l\left(j\right)$ are pixels' luminance values. $\sigma_{i}^{2}$ is the luminance variance of neighborhood pixel set of $i$.

\section{Experimental analysis}
For our model, we collect training set from Internet. Because the preparation of paired images is extremely labor-intensive, we sampled a small volume of data. In our experiments, we collect only 1000 pairs of training data. These images (image size is around 400$\times$600 pixels) contain various scenes and objects, such as flowers, trees, animals, buildings and so on. We then use AP algorithm~\cite{Xiao:2007} to cluster these images based on their global features to auto construct several categories of the images. After applying the training algorithm to each image category, we can obtain the parameters for every category of the images. The cluster center on GLCM based global features and the corresponding parameters for every category after training are shown in table~\ref{tab:trainResults}.

\begin{table*}[htp]
	\caption{The cluster center of the global features and the corresponding parameters after training for each category.}
	\label{tab:trainResults}
	\centering
	\begin{tabular}{p{1.0cm}<{\centering}|p{2.0cm}<{\centering}|p{1.0cm}<{\centering}|p{1.0cm}<{\centering}|p{1.0cm}<{\centering}|p{1.0cm}<{\centering}|p{1.0cm}<{\centering}|p{1.0cm}<{\centering}|p{1.0cm}<{\centering}|p{1.0cm}<{\centering}}
	\hline
	\multirow{2}{*}{Category} & \multirow{2}{*}{Sample numbers} & \multicolumn{4}{c}{Cluster Center} & \multicolumn{4}{|c}{Optimal Weight Parameters}\\
    \cline{3-10}
    & & ASM & CON & CORRLN & ENT & Intensity & STD & SURF & Gabor\\
    \hline
    1 & 107 & 0.9652 & 0.7745 & 0.8544 & 0.0793 & 0.1007 & 0.1734 & 0.4509 & 0.2750\\
    \hline
    2 & 213 & 0.1720 & 0.9331 & 0.9626 & 0.2004 & 0.1874 & 0.1765 & 0.3971 & 0.2391\\
    \hline
    3 & 244 & 0.6516 & 0.8195 & 0.8939 & 0.0961 & 0.1217 & 0.1712 & 0.4484 & 0.2587\\
    \hline
    4 & 241 & 0.3052 & 0.8952 & 0.9509 & 0.1433 & 0.1555 & 0.1978 & 0.3819 & 0.2647\\
    \hline
    5 & 195 & 0.4556 & 0.8593 & 0.9241 & 0.1155 & 0.1376 & 0.1749 & 0.4127 & 0.2748\\
    \hline
	\end{tabular}
\end{table*}

 {To evaluate our approach, we collect another 500 pairs of color images to construct the test set manually. Each pair contains two color images whose color style are similiar, we transform one of the image in each pair into the gray scale image to construct the target image, and the other one is regarded as the reference color image. All the transformed grayscale target image and the correpsonding reference color image pairs will consist of the test set, and thus we can have the ground truth for each target image in the test set.} In the following experiment, we employ six methods, i.e., Welsh et al.~\cite{Welsh:2002}, Irony et al.~\cite{Irony:2005}, Gupta et al.~\cite{Gupta:2012}, Bugeau et al.~\cite{Bugeau:2014}, Zhang et al.~\cite{zhang2016colorful}, He et al.~\cite{he2018deep} and our proposed approach to colorize all these gray images in the test set. We keep all the parameters the same as the authors' of all those six other state-of-the-art methods in our colorization experiments.

Specifically, as for Welsh et al.'s method, we use rectangular swatches to match areas between the grayscale input image and the reference color image. As for Irony etl al.'s method, we use meanshift algorithm to segment the reference image, and the segmented image is then used as input with the grayscale image. As for Gupta et al.'s method, the super-pixel size is around 40 pixels, and the weights of intensity, standard deviation, Gabor and SURF are set to be 0.2, 0.1, 0.2, 0.5 respectively in the process of feature matching. We use Bugeau et al.'s open source code in all the following experiments.

As for Zhang et al.'s method, They propose four models which are trained on 1.3M images from ImageNet. We use the model trained without class rebalancing to test the 500 input images because this model will provide ``safer" colorizations.

To further compare with the learning-based method, we experiment with a method called ``ZhangS", which uses Zhang et al.'s network architecture (Zhang et al.'s class method) to perform the training process with our 2500 collected images (including 2000 training samples and 500 reference color images in the test set) as the training set. The training images were warped and downsampled to 256x256 for the input requirement of the network. The network was trained from scratch with k-means initialization~\cite{krahenbuhl2015data}, using the ADAM solver for 80k iterations ($\beta_1=0.9$, $\beta_2=0.99$, weight decay = 0.001). Initial learning rate was $3.16\times10^{-5}$ and dropped by multiplying it by a factor of $\gamma = 0.316$ every 20k iterations. We kept the mini batch size 40 the same as Zhang et al.'s settings. The training process will converge to a stable loss after ~80k iterations, so we saved the model at 80k iterations and used it to test the 500 input images.

We also experiment with a ``ZhangF'' method fine-tuned from Zhang et al.'s model, with our 2500 images (including 2000 training samples and 500 reference color images in the test set) as training set. All the settings in ``ZhangF" are the same as ``ZhangS" except for the network initialization and the value of base learning rate. We restored Zhang et al.'s class model (trained on 1.3M images from ImageNet for approximately 450k iterations without class rebalancing) for the network initialization. And set the value of base learning rate slightly smaller ($10^{-5}$). We trained for ~50k iterations when the loss plateaued at a low level. Thus we use the model at 50k iterations to test the 500 input images.

As for He's method, we use their model trained on approximately 0.7M image pairs to colorize our 500 test images.

\subsection{Colorization Results}
\begin{figure*}[htp]
	\centering
	\includegraphics[scale=0.68]{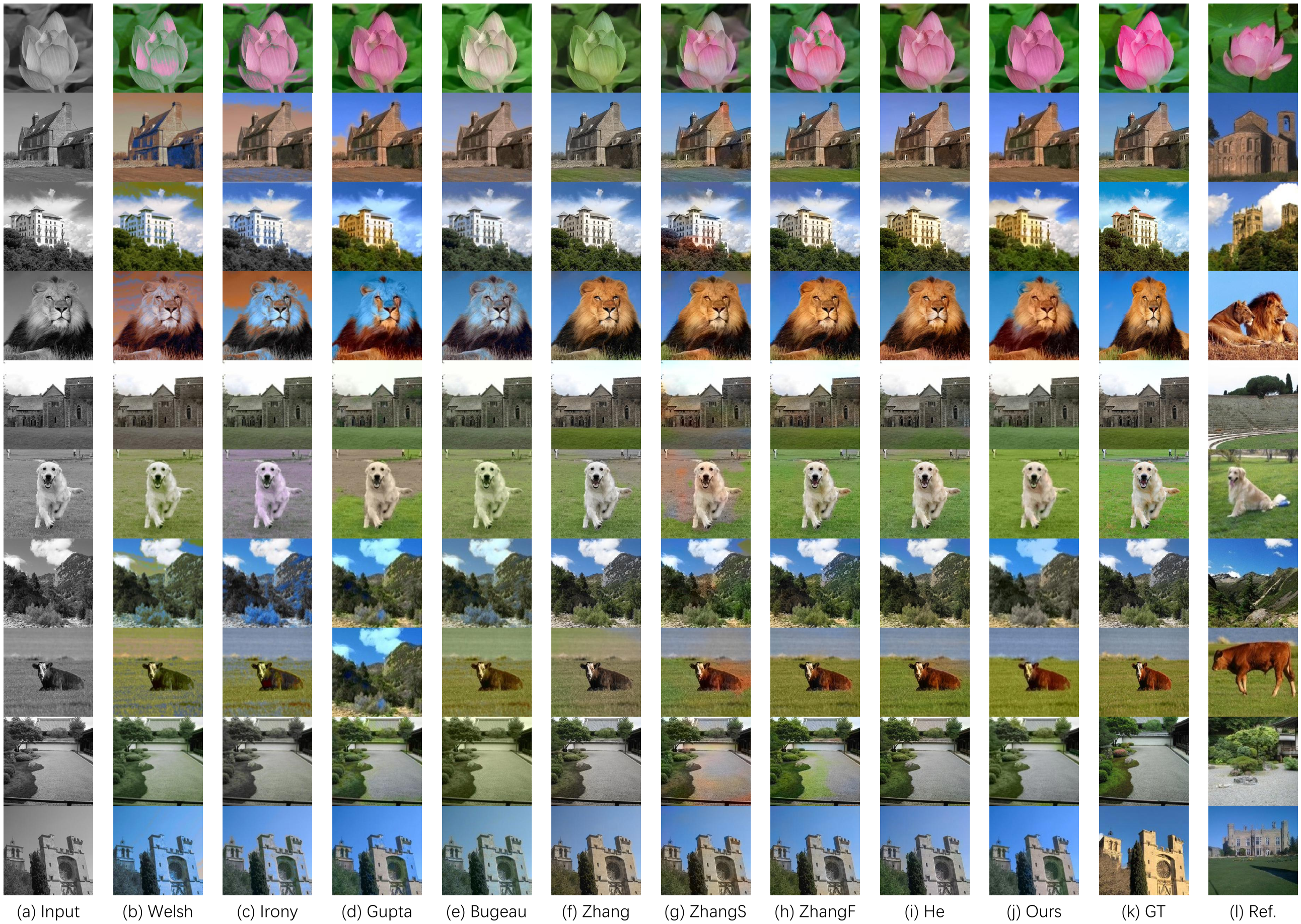}
	\caption{Comparison with the state-of-the-art colorization methods. The first column shows the grayscale input images. The second to tenth columns show the colorization results obtained by nine difference methods. The penultimate column is the reference color images. And the last column is the ground truths.}
\label{fig:goodResults}
\end{figure*}

Parts of the colorization results are shown in Fig.~\ref{fig:goodResults}. The first column is the grayscale input images. The second to sixth columns, together with the tenth column, are colorized images obtained from six different methods with the reference images as shown in the last column. The seventh to ninth columns are colorized images obtained from learning-based methods.

From the colorization example of lotus we can see that our method can distinguish lotus in Fig.~\ref{fig:goodResults} and surrounding leaves distinctly, and colorize lotus with pink, leaves with green. While Welsh et al.'s method~\cite{Welsh:2002} colorizes part of the lotus into green, which may be caused by the reason that it distributes colors in the level of pixel without considering the neighborhood spacial information,  and thus lead to discontinuity in the color space. As both the target image and reference image in the first row are hard to differ the border between the lotus and leaves, Irony et al.'s method~\cite{Irony:2005} also colorizes part of the leaves into pink due to the unreasonable segmentation in the reference image. When considering the results of Gupta et al.'s method~\cite{Gupta:2012} and Bugeau et al.'s method~\cite{Bugeau:2014} in the lotus, they both achieve quite worse results compared to other methods, the reason may be that they all make the assumption that pixels with similar intensities or similar neighborhood should have similar colors. While the gra scale input image of lotus have similar intensities and textures for the lotus and leaves, but the colors for lotus and leaves are obviously different. This condition can also be notable in the forth row of the image of lion, where the intensity of the lion's hair around the face is similar to the sky's, and then both Gupta and Bugeau make the wrong colorization.

\begin{figure*}
	\centering
	\includegraphics[scale=0.9]{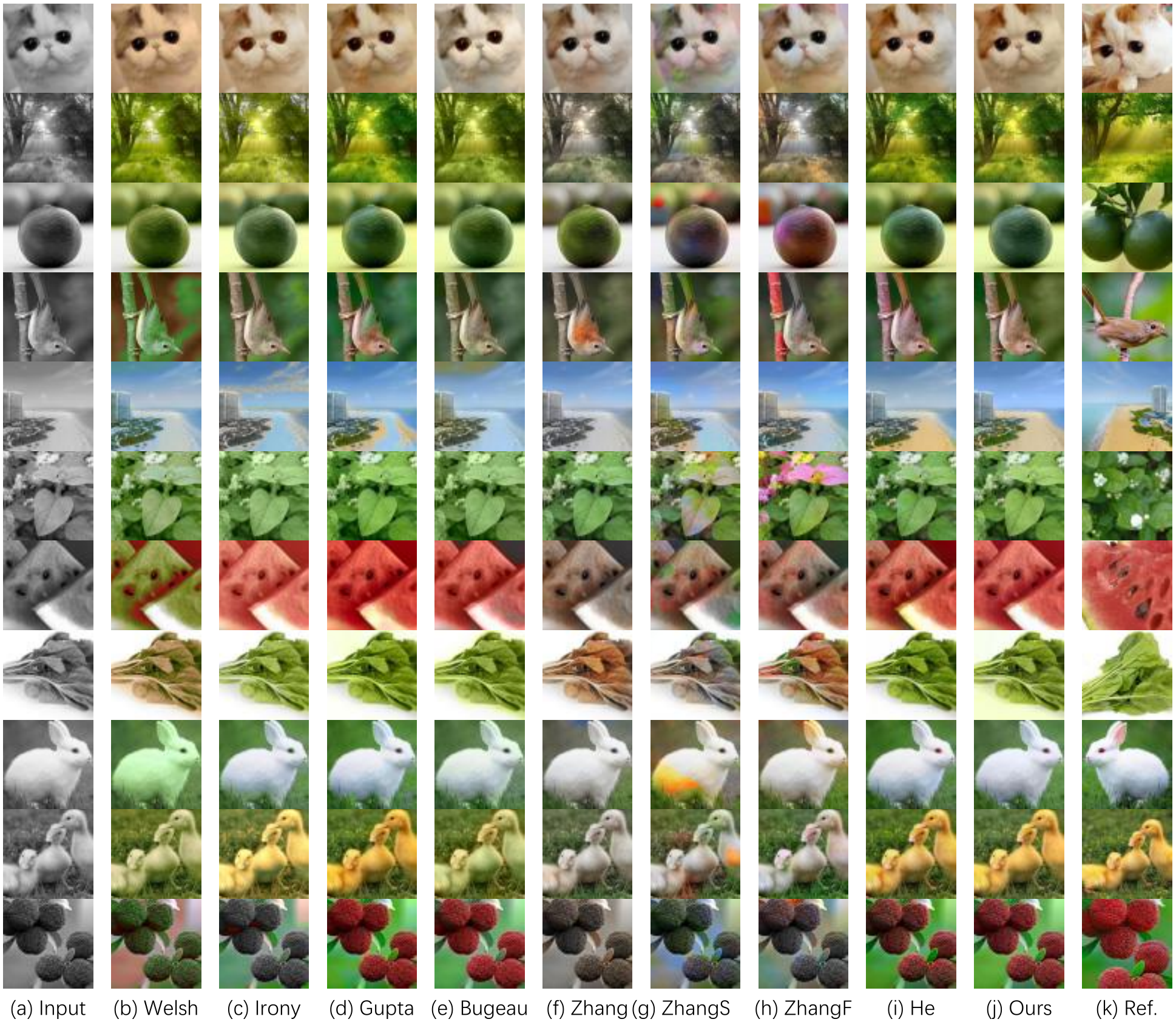}
	\caption{More visualization results used in the user study experiment.}
\label{fig:moreResult}
\end{figure*}

Zhang et al.'s method~\cite{zhang2016colorful} colorizes the whole lotus with green. That's because there are few scenes similar with lotus in the training data, so that the trained model can not use history information to colorize the lotus with reasonable color. More results are shown in Fig.~\ref{fig:moreResult}.

``ZhangS" performs better than Zhang in lotus example. This is because ``ZhangS" is trained with our collected 2500 images, where contain similar samples with the input grayscale image. However, the edges of the lotus are still colorized with green, and some parts of the background are colorized with pink. In the other hand, Zhang et al.'s method performs much better than ``ZhangS" in Fig.~\ref{fig:moreResult}, which is because there are lots of similar images in ImageNet with these target images.

``ZhangF" has better performance than the previous two learning-based methods. Because ``ZhangF" is fine-tuned from 
 model with our 2500 images as training data set. So it not only maintains the color modes from large-scale images of ImageNet, but also learns the new modes from the 2500 images. However, some parts of the lotus are still colorized with green. In Fig.~\ref{fig:moreResult}, ``ZhangF" colorizes the whole images with a color deviation, while our method will pull the color offset back into the reasonable position with the given reference image. Meanwhile, by using the different structure parameters in colorization, our method performs quite good in these examples.

He et al.'s method~\cite{he2018deep} has better performance than pure learning-based methods or pure exemplar-based methods above all the test images. But it still colorize some leaves in the lotus image (row one) with wrong colors. And our method performs slightly worse than He et al.'s only in the lion image (row four). In the next section, we will further compare our method with those state-of-the-art methods in detailed quantitative analysis, and show the comparability and superior performance of our method.

Although our approach also assumes that pixels with similar intensities or similar neighborhood should have similar colors, the pre-learned weight parameters for matching in our learning framework have encoded the priori knowledge that those patches with similar intensities or textures should not be matched. That is why our method can achieve much better results in both images of the lotus and lion.

\subsection{User Study}
We also carry out a user study to further compare our colorization method with other state-of-the-art methods. We invite 30 volunteers to evaluate the colorization results of those nine methods. Specifically, we show each volunteer a set of ten images at a time, they are required to select the best three reasonable colorized images within ten seconds. We totally provide 40 experiment sets for each volunteer. In order to prevent the volunteer's mindset, we randomly disrupted the order of nine methods in each experiment set.

We put forward a token $I_{ijk}$ to represent the volunteer's choice. If volunteer $i$ chooses the image colorized by method $k$ in the $j$th image set, and then $I_{ijk}=1$, otherwise $I_{ijk}=0$. We calculate the average percentage of each method in this experiment with:
\begin{equation}
{p_k} = \frac{{\sum\limits_{i = 1}^{30} {\sum\limits_{j = 1}^{40} {{I_{ijk}}} } }}{{30 \times 40 \times 3}}.
\end{equation}

From the results (shown in table~\ref{tab:userStudy}) we can see, volunteers choose our method and He et al.'s method (slightly lower than our method) as the most reasonable colorization methods. This compares favorably to the methods using learning-based algorithms and other exemplar-based methods.

\begin{table*}[htp]
\centering
\caption{{Uer study results of our method compared with eight state-of-the-art methods.}}\label{tab:userStudy}
  \begin{tabular}{p{1.8cm}<{\centering}|p{1.0cm}<{\centering}|p{1.0cm}<{\centering}|p{1.0cm}<{\centering}|p{1.0cm}<{\centering}|p{1.0cm}<{\centering}|p{1.0cm}<{\centering}|p{1.0cm}<{\centering}|p{1.0cm}<{\centering}|p{1.0cm}<{\centering}}
    \hline
    Method & Welsh & Irony & Gupta & Bugeau & Ours & Zhang & ZhangS & ZhangF & He\\
    \hline
    Percentage(\%) & 2.76 & 4.49 & 11.92 & 7.60 & 18.94 & 11.87 & 8.09 & 15.43 & 18.90\\
    \hline
  \end{tabular}
\end{table*}

\subsection{Evaluation Experiments on the Weight Parameters}
The core idea of our approach is to learn proper weight parameters for those four features, which are used to match the corresponding super-pixel from the reference color image. Thus we also design a quantitative experiment to evaluate the effectiveness of our learning framework. In this experiment, we randomly generate 20 sets of weight parameters, which are shown in table~\ref{tab:weightsParameters}, and use them to colorize the 500 grayscale input images without selecting the corresponding trained weight parameters via the input target image's GLCM-based global features. Since we have the ground truth of those 500 test images, we can calculate the average error and the average error rate of each method on the test set as indicators to evaluate the performance of the parameters.

\begin{table*}[t]
\centering
\caption{The randomly generated weight parameters used in the experiments.}~\label{tab:weightsParameters}
  \begin{tabular}{p{0.6cm}<{\centering}|p{1.2cm}<{\centering}|p{1.2cm}<{\centering}|p{1.2cm}<{\centering}|p{1.2cm}<{\centering}|p{0.6cm}<{\centering}|p{1.2cm}<{\centering}|p{1.2cm}<{\centering}|p{1.2cm}<{\centering}|p{1.2cm}<{\centering}}
    \hline
    & \multicolumn{4}{c|}{Random Parameters} & & \multicolumn{4}{c}{Random Parameters} \\
    \hline
    1 & 0.2435 & 0.0547 & 0.5519 & 0.1499 & 11 & 0.0615 & 0.3931 & 0.3121 & 0.2332 \\
    \hline
    2 & 0.4524 & 0.1329 & 0.0608 & 0.3539 & 12 & 0.2177 & 0.0940 & 0.4995 &	0.1889 \\
    \hline
    3 & 0.0698 & 0.1621 & 0.0718 & 0.6963 & 13 & 0.1655 & 0.2596 & 0.3392 &	0.2357 \\
    \hline
    4 & 0.3669 & 0.0287 & 0.2922 & 0.3122 & 14 & 0.2259 & 0.2603 & 0.2480 &	0.2658 \\
    \hline
    5 & 0.5580 & 0.2879 & 0.0575 & 0.0967 & 15 & 0.0187 & 0.1217 & 0.4827 &	0.3769 \\
    \hline
    6 & 0.2542 & 0.0761 & 0.2395 & 0.4303 & 16 & 0.1199 & 0.5187 & 0.1006 &	0.2607 \\
    \hline
    7 & 0.0316 & 0.2830 & 0.3398 & 0.3457 & 17 & 0.3619 & 0.1025 & 0.1261 &	0.4095 \\
    \hline
    8 & 0.1799 & 0.3222 & 0.2607 & 0.2372 & 18 & 0.2989 & 0.2125 & 0.4348 &	0.0537 \\
    \hline
    9 & 0.0307 & 0.1790 & 0.4632 & 0.3271 & 19 & 0.3024 & 0.2859 & 0.2402 &	0.1715 \\
    \hline
    10 & 0.4589 & 0.2928 & 0.0879 & 0.1604 & 20 & 0.2465 & 0.0823 &	0.2871 & 0.3840 \\
    \hline
  \end{tabular}
\end{table*}

The average error of an image on color matching is:
\begin{equation}
M =\frac{\sum_{t_{i}}\left\|C\left(\check{t}_{i}\right)-C\left(t_{i}\right)\right\|}{S},
\end{equation}
where $S$ is the number of super-pixels in the image. Then the average error on a test set can be calculated as:
\begin{equation}
avgE=\frac{1}{N}\sum_{N}M_{j} \left(j=1,2,3,\cdots,N\right),
\end{equation}
where $N$ is the image number of the test set.
We define the error rate of an image as follow:
\begin{equation}
A_{j}=\frac{\sum_{t_{i}\in \left\{\left\|C\left(\check{t}_{i}\right)-C\left(t_{i}\right)\right\|>\theta\right\}}1}{S} \left(i=1,2,3,\cdots,S\right),
\end{equation}
where $\theta$ is the color error threshold. Here we set $\theta=3\pi$ empirically.
Then the average error rate of the test set is:
\begin{equation}
A=\frac{1}{N}\sum_{N}A_{j} \left(j=1,2,3,\cdots,N\right).
\end{equation}

\begin{figure}[htp]
  \centering
  \includegraphics[scale=0.58]{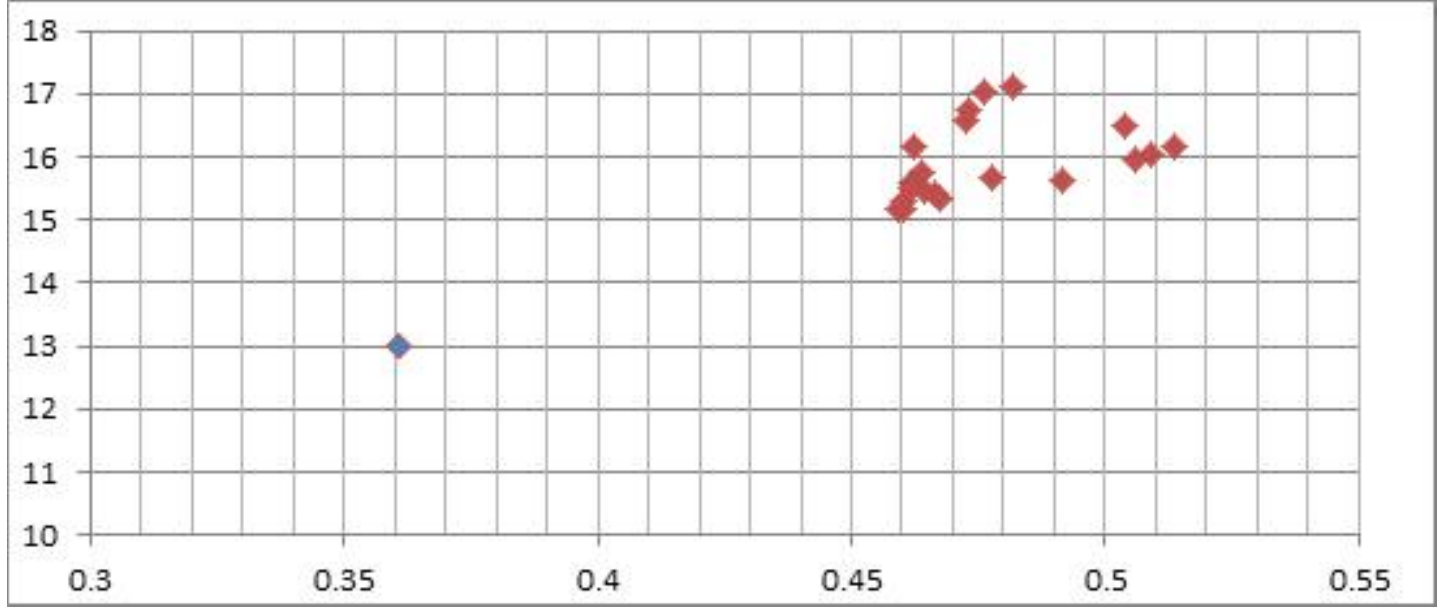}
  \caption{Comparison results of the average error rate (x-coordinate) and the average error (y-coordinate) on the test image set. The results of those 20 randomly generated weight parameters are denoted with red points, and the result of our learned parameters is denoted with blue point.}~\label{fig:20rw}
\end{figure}

We calculate the average error and the average error rate on the test set with the 20 random weight parameters, and meanwhile, the average error and the average error rate on the same test set of our approach are also calculated. The statistical results are shown in Fig.~\ref{fig:20rw}, where $X$ axis represents average error rate, $Y$ axis represents average error, we use red points to represent the calculation results obtained by those randomly generated weight parameters and the blue point to represent the result of our learned weight parameters. Obviously, the learned parameters are much better than random parameters on both the average error and the average error rate.

\subsection{Quantitative Experiments on Color Difference}
We also design experiments to evaluate the color difference between the colorized image and its corresponding ground truth.
As the color difference between two images is hard to judge by vision directly, we adopt the evaluation standards proposed by CIE (International Commission on Illumination)~\footnote{http://www.cie.co.at/} to quantitatively analyze the color difference between the colorized image and its ground truth. 

There are several reasons why we choose color difference metric to evaluate the colorization performance. Firstly, in our experiment, we manually pick the reference color images whose color style are reasonable to the target grayscale images. Secondly, as for colorization methods with one grayscale image and one reference image as input, such as~\cite{he2018deep}, the color difference metric can be used to evaluate whose colorization performance is closer to the GT. Finally, as for the colorization methods with only one grayscale image as input, such as~\cite{zhang2016colorful}, the colorized image has nothing to do with the reference image, so the color difference metric is a relatively fair evaluation standard.

CIE proposed several standard methods to evaluate the color difference, such as CIE76, CIE94, CIEDE2000~\cite{sharma2005ciede2000}. CIE76 calculated the Euclidean distance between two colors in $l\alpha\beta$ color space directly, but the effect is not as perceptually uniform as intended. CIE94 refined the color difference calculation formulas for the discontinuity. CIEDE2000 resolved the perceptual uniformity issue by adding five corrections, which formed the color difference appraisal system in solving perceptual uniformity till now. Our experiment is based on CIEDE2000 to judge the colorization effect.

For the corresponding pixel-pairs of the colorized image and the ground true image,
the color difference defined in CIEDE2000 is a function on the pixel-pairs' $l\alpha\beta$ values, and denoted as $\Delta E_{00}(L_1^\ast,a_1^\ast,b_1^\ast,L_2^\ast,a_2^\ast,b_2^\ast)$. The calculation process of $\Delta E_{00}$ is given in ~\cite{sharma2005ciede2000} .

We define an evaluation index for pixel $p$ in colorized image $I$ as
\begin{equation}
e_{p} = \left\{ \begin{array}{ll}
1 & \textrm{$\Delta E_{00}(p,\check{p})>$ T}\\
0 & \textrm{$\Delta E_{00}(p,\check{p}) \leq$ T}
\end{array} \right.
\end{equation}
$\Delta E_{00}(p,\check{p})$ represents the color difference of pixel-pairs between the colorized image and the ground true image. We decide that the colorized color in a pixel is unsuitable when the difference is bigger than threshold T. Thus the pixel is defined as a bad point. The colorization error evaluation function of image $I$ is defined as
\begin{equation}
errorRate =
\frac{\sum_{p\in I}e_p}{N_I}
\end{equation}
$N_I$ is the number of pixels in image $I$. Then the Mean Error Rate is computed over the entire test set.

\begin{figure}[htp]
    \centering
    \includegraphics[scale=0.48]{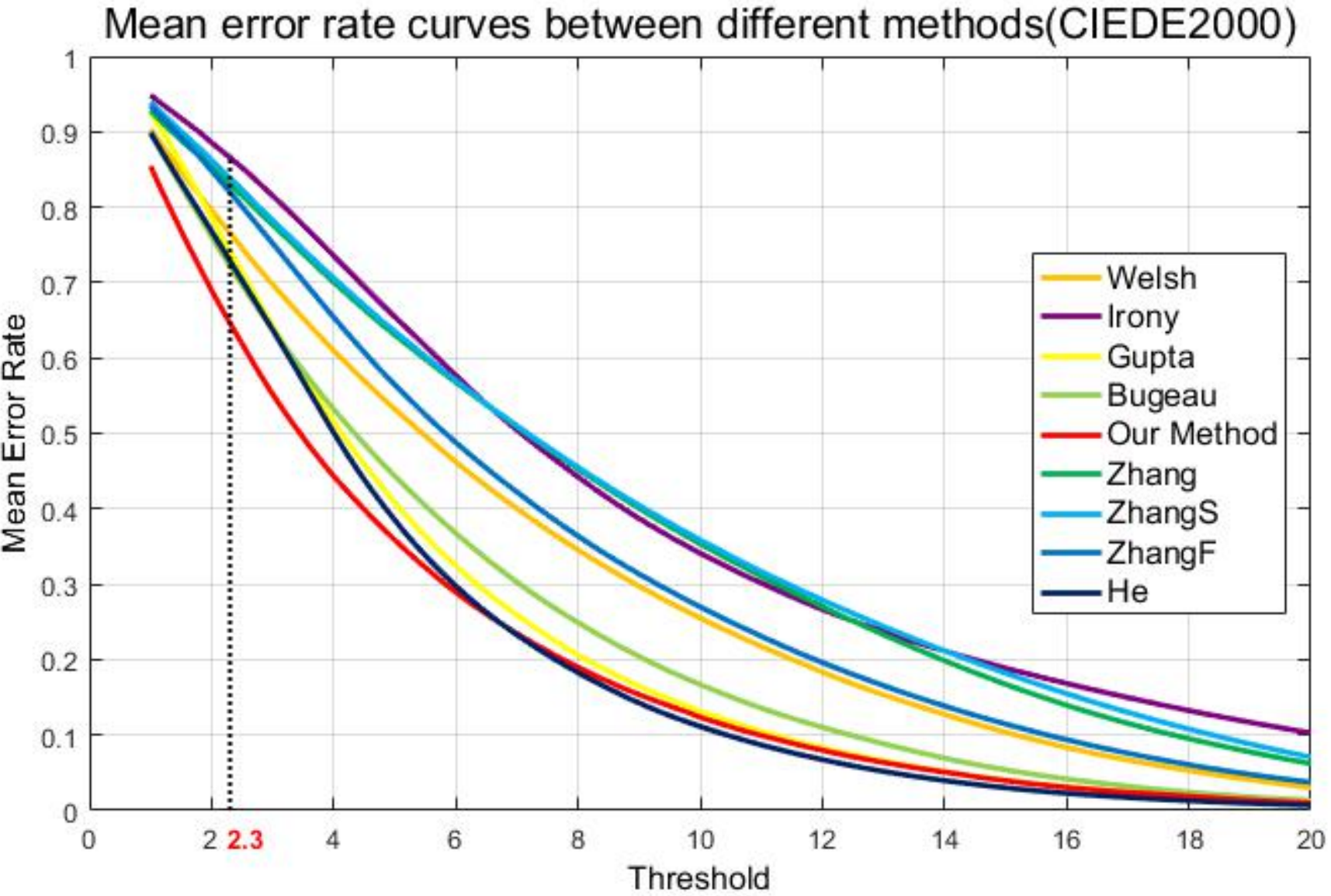}
    \caption{Mean error rate curves of different methods.}
    \label{fig:MER}
\end{figure}

We use the results obtained by the eight colorization methods mentioned above and our proposed method to calculate the Mean Error Ratio, shown in Fig. ~\ref{fig:MER}.

When the color difference threshold is large enough, which means a high color difference between colorized image and the original image is tolerated, the result is that even though the output image is poorly colorized, Mean Error Rate is still at a low level. So the Mean Error Rate of different colorization methods are similarly close to zero in high threshold zone. The result in Fig.~\ref{fig:MER} shows that our method can always achieve the lowest error rate in the experiment (slightly higher than He et al.'s method when threshold is large).

There are also researches show that the values of $\Delta E$ are directly related to the human's awareness of the color difference, which means there is a value of $\Delta E$ approaching to the JND(Just Noticeable Difference)~\cite{bala2003digital} of human. Some studies assessed a value of 2.3 on $\Delta E$~\cite{mahy1994evaluation} can approach to the critical color difference just to be noticed. when $\Delta E$ is bigger than 2.3, our human eyes can distinguish the color difference easily. We plot a vertical line at 2.3 on the horizontal axis in Fig.~\ref{fig:MER}, the Mean Error Rate of our method is around 0.65 at the threshold of 2.3 (JND threshold), while Bugeau's is 0.73 and Irony goes up to 0.87. That show our method has a significant advantage when consider the color difference at JND threshold~\footnote{As the choosing of the reference images may lead to different error rates, the relative values of every methods' error rates on the same reference image are more indicative.}.

The results in Fig.~\ref{fig:MER} also shows that the mean error rates of learning based methods are much larger than our method, and even larger than Bugeau et al.'s method and Gupta et al.'s method, which also indicates that those learning based methods may have larger color bias for the whole image as they colorize grayscale images only by the encoder learned from large-scale image set while not employ the high confidenced knowledge provided by the referenced color images.

\subsection{Experimental evluation on the four features}
In this section, we will carry out experiments to evalute the importance of those four features, i.e. intensity $w_1$, standard deviation $w_2$, Gabor feature $w_3$, and SURF descriptor $w_4$ respect to the colorization. 

In order to measure the importance of the individual feature, we will vary the single weight while keep the other weights with the same proportion as the weights learnt by our training method. Suppose the weight of a catgory of images leart by our framework is $W=(w_1, w_2, w_3, w_4)$, we will construct four weight sets, $\{W^*\}_{w_i}$, by varying the value of $w_i(i=1,2,3,4)$ from 0 to 1, while the other three weights are calculated as following:

\begin{equation}
w_j^* = \frac{w_j}{1-w_i}, j\in {\{1,2,3,4\}}, j \not= i
\end{equation}

For every category of images, we vary the $w_i$ by arithmetic progression from 0 to 1, with common difference of 0.1. Then it will generate four weight sets, $\{W^*\}_{w_i}$, corresponding to the $W$~\footnote{Each feature weigh $w_i$ will be varied to obtain 11 $w^*_i$, and there are four features and 44 $W^*$ in total.}. In our experiments, we choose 500 test images that belongs to different categories, for the learnt weight $W$ of each category, we will generate its corresponding  $\{W^*\}_{w_i}$. We then use the $W$ and its corresponding  $\{W^*\}_{w_i}$ to colorize those images containing in the category respect to $W$. 

\begin{figure}[htp]
    \centering
    \includegraphics[scale=0.48]{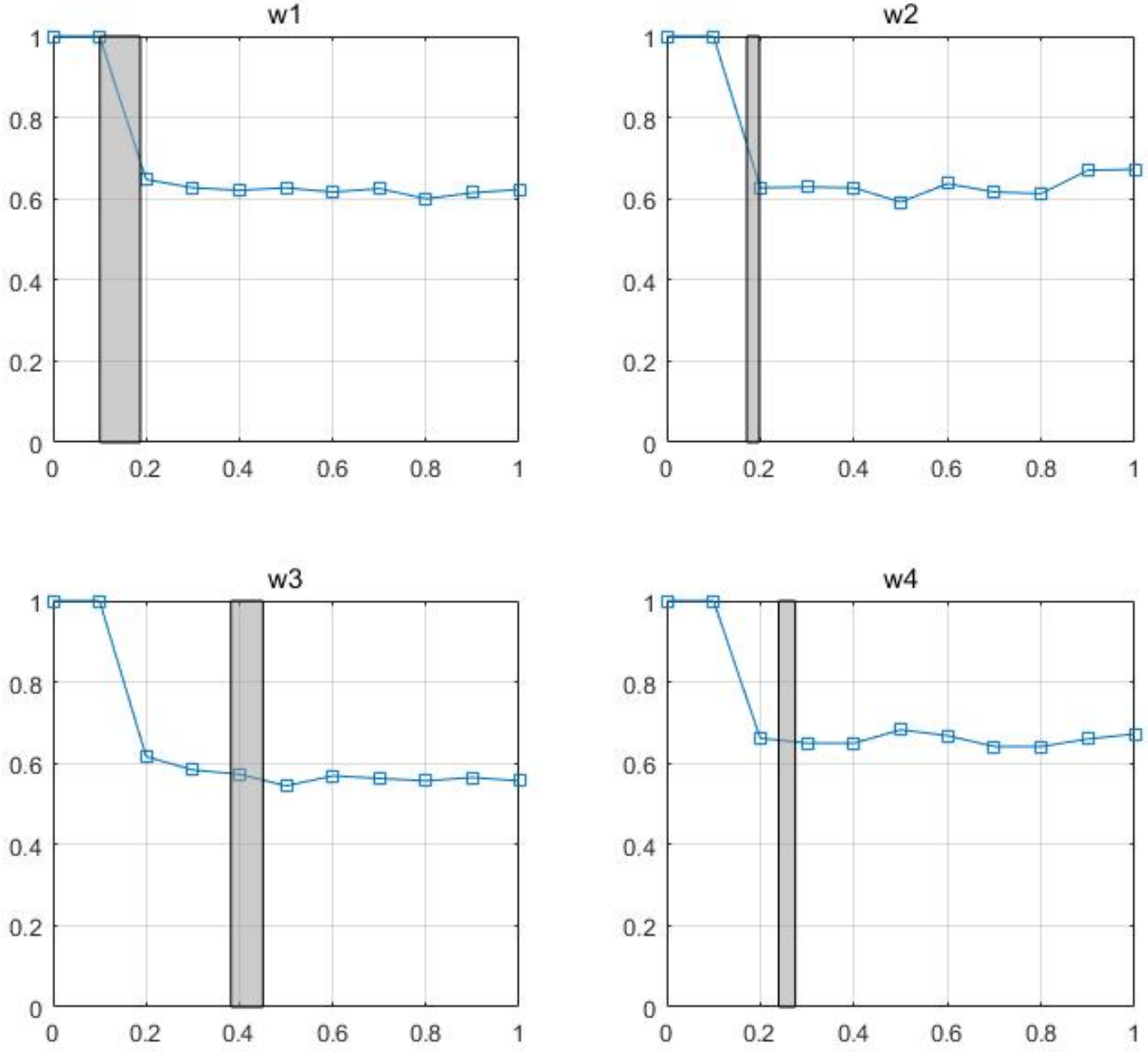}
    \caption{Colorization performance of each feature under varied weights. The x-axis is the weight variation range of the corresponding feature ($w_i$). Each sub-figure shows the performance addressing on the individual feature. y-label is the ratio of the images that our learned weight $W$ can achieve less $JND_{2.3}$ error than  $\{W^*\}_{w_i}$. The gray vertical bar shows the variation range of our learned weights under those 500 test images.}
    \label{fig:experiment-new-w1}
\end{figure}

After obtaining the colorized images, we calculate the quantitative error based on JND at the threshold of 2.3. For each feature of the weight $w_i$, we will consider the ratio of the images whose errors obtained by our learnt  weight $W$ are less than the errors obtained by $\{W^*\}_{w_i}$ on those 500 test images. The results for those four features ($w_1, w_2, w_3, w_4$) are shown in Fig.~\ref{fig:experiment-new-w1}. 

The results in Fig.~\ref{fig:experiment-new-w1} show that all the y-labels of the sub-figures will appraoch to 1 when the corresponding features' weights are equal to 0, which means the performance of $W$ will be totally better than $\{W^*\}_{w_i}$ when $w^*_i = 0$, and it also indicates that these four features are all necessary to the colorization. When considering the variation ranges of the learnt (optimal) weights under the performance curves, we can find that the learnt ranges of $w_1$ and $w_2$ will lie on the steep sections of their curves, while $w_3$ and $w_4$ will lie on the flat sections of their curves, which indicates that $w_1, w_2$ may be more sensitve to the performance compared with $w_3, w_4$, which means small changes on $w_1, w_2$ will lead to big performance changes.

\subsection{Experimental analyse of generalization ability}
In this section, we will show the generalization ability of our framework. Firstly, We colorize the input grayscale images with the weights learnt by five categories, these categories are automated clustered in the trainined component. And then we visualize the images in the five categories to find commonalities. Detailed experiment settings are shown below.

\begin{figure*}[htp]
	\centering
	\includegraphics[scale=0.9]{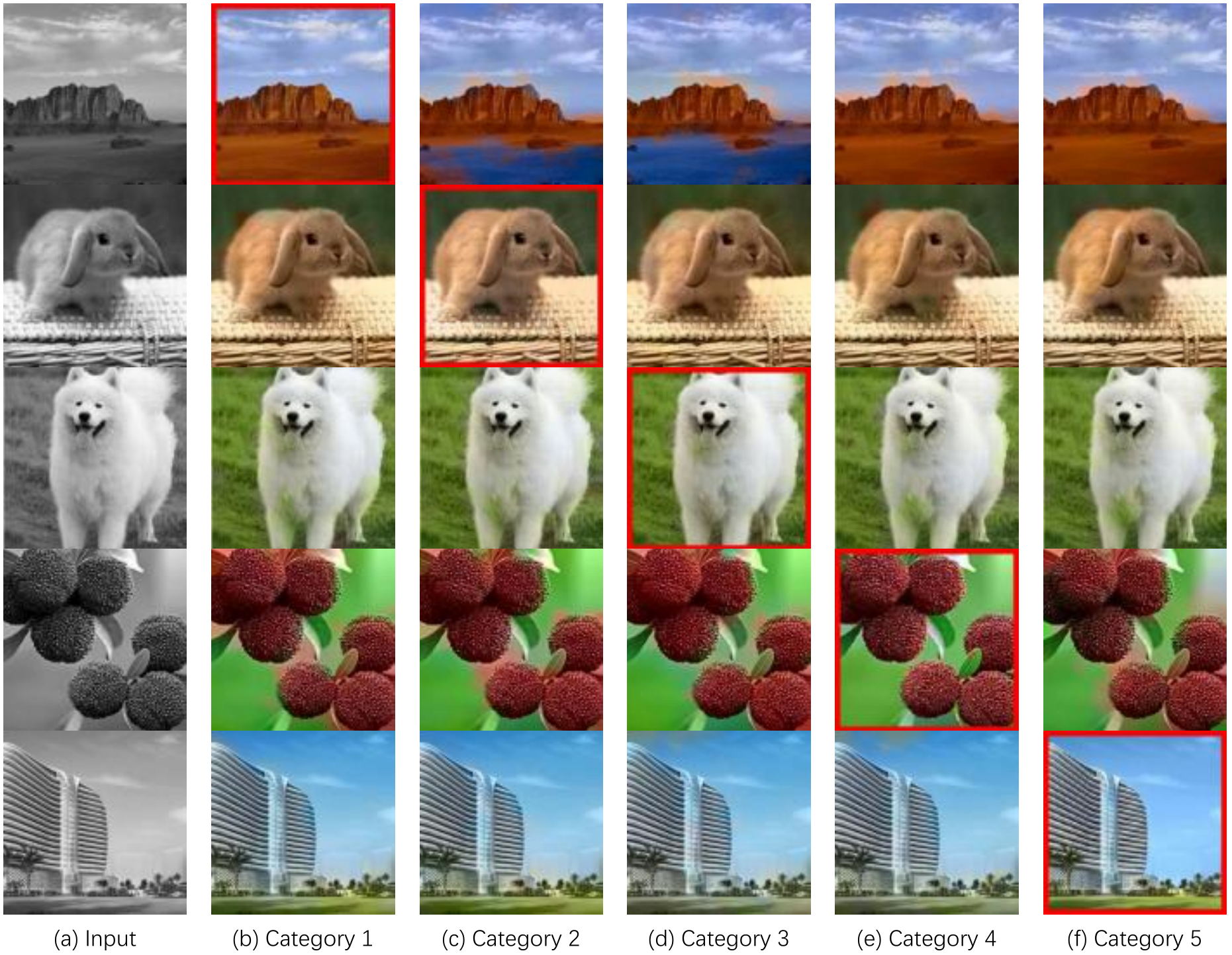}
	\caption{Colorization results with weights learnt by five categories. Column one is the target grayscale image. Column two to six are images colorized with the weights from five different categories. The image colorized by the most appropriate weights is circled with a red box.}
    \label{fig:colorization-results-5-categories}
\end{figure*}

Firstly, we colorize the input grayscale images with the weights learnt by the five categories, which are learnt by our 1000 pairs training data set. The detailed information is shown in table~\ref{tab:trainResults}. As for each input image, we will classify it into one category by global feature vector, and the weight of this category is most appropriate for colorization process. The colorization results are shown in Fig.~\ref{fig:colorization-results-5-categories}, where column one is the target grayscale image. Column two to six are images colorized with the weights from five different categories. The image colorized by the most appropriate weights (which means the input gray image is closet to that category)  is circled with a red box. We can see that colorized image with red box has the best performance compared with the other four colorized images most of the time, and the other four categories of feature vectors can also achieve reasonable colorization results.

\begin{figure*}[htp]
	\centering
	\includegraphics[scale=0.9]{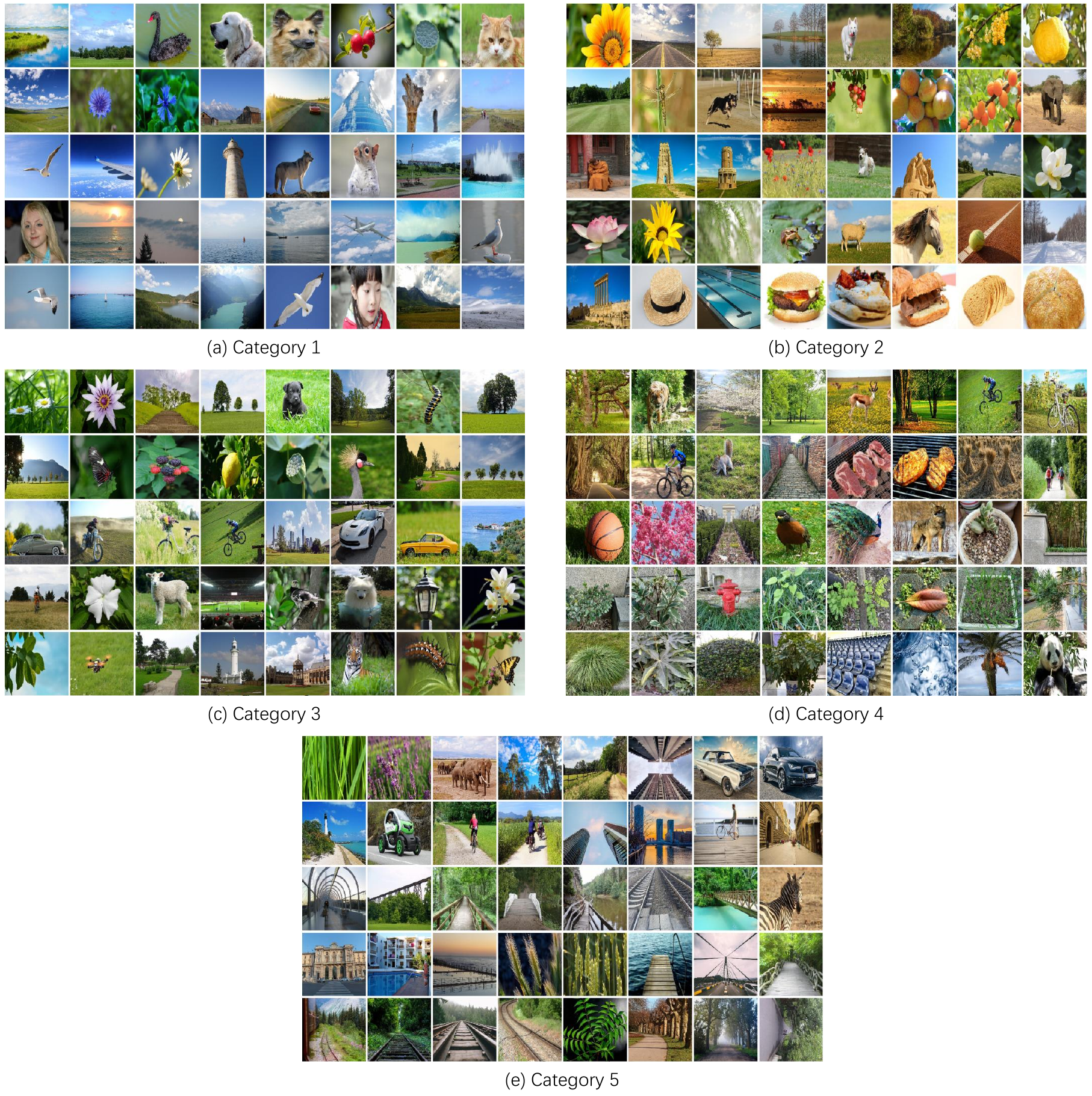}
	\caption{Visualization of the five categories in training data set. Each category is sampled with 40 images. Note that these color images are corresponding to the grayscale images used in the process of clustering. And We find that different category has different features ignore the color. e.g. Catetory 1 is monotonous, Catetory 2 is delicate, Catetory 3 is relatively rich in structure. While Category 4 has more burrs. The lines in Category 5 are more prominent.}
    \label{fig:visualize-5-categories}
\end{figure*}

We also visualize each category in the training data set with 40 sampled images (as shown in Fig.~\ref{fig:visualize-5-categories}). We can see that different category has different features. A given image will be always classified into one category by its global features, since there is no clear classification boundary between the five categories and out-of-category.
So that once the image is properly classified into one category, our system will colorize it with the most appropriate color.

\subsection{Experiments on different reference images}
To further evluate our method's performance under different reference images, we also carry out experiments to colorize the gray images with varied reference images.

\begin{figure*}[htp]
	\centering
	\includegraphics[scale=1.4]{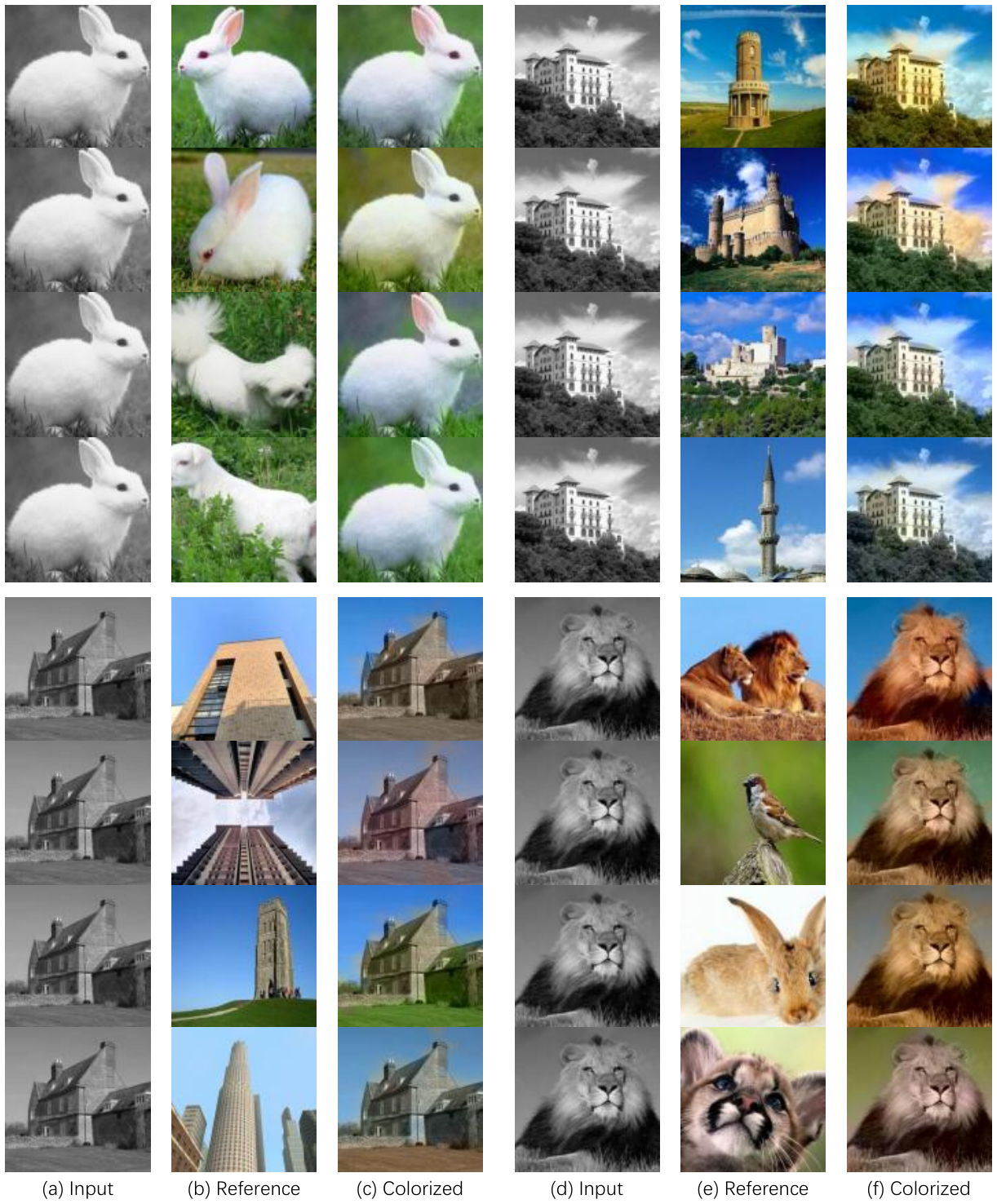}
	\caption{Colorization results with different reference images. The first and fourth columns are four grayscale images. We select different reference color images for each target image, which are shown in second and fifth columns. The third and last columns are colorized images for each target image by using the corresponding reference images.}
    \label{fig:different-refer1}
\end{figure*}

The selection of the reference image is the key for the performance of the proposed model. In real application, however, the color style similarity is subjective: (1) there are no guidelines or criteria to restrict the user to pick the reference image when the model is deployed in real applications. So we think the best way we can approximate the input data space of the user, is to collect data space from also human users, instead of some math criteria; (2) in our task, the freedom of picking the reference image also eliminates the assumption in some objective image colorization model that the grayscale image to color image is one-to-one mapping as in~\cite{zhang2016colorful}. As shown in Fig.~\ref{fig:different-refer1}, our model can generate one-to-many colorization results when the color style of the picked reference image is considered reasonable.

The results show that our method is robust to output various color styles for the same input image with different input reference images, which indicates that our approach can successfully learn the colorization process while decoupling the colorization styles as the input of reference images.

\subsection{Summary of Advantages and Limitations}
The main advantage of our method is that it is able to combine the advantage of learning based methods that can adapt to various structural composition of images with different scenes or objects and the advantage of exemplar-based method that can keep a small global color bias during the colorization. As the training color images can be quite easily obtained and their ground truths are also known, our method can almost handle all kinds of images in practice. That is why our method can perform better in the previous experiments. The second advantage is the introduction of the global feature of images, which provides a measure metric for the composition of the images, and thus we can adopt the idea of ``divide-and-conquer'' to cluster those images with similar compositions and learn different optimal parameters for every clusters. This will help our approach solve the difficulty caused by various compositions of the target images, and it will also support parallel learning for very large-scale image set.

As our method can be regarded as a hybrid of the learning-based method and exemplar-based method, on the contrast side, the learning-based method can also be regarded as a special case of the exemplar-based method, which encodes the reference images into its learned models. That will explain why the learning-based methods are highly sensitive to their training data set. In order to adapt to various structural composition of images, the learning-based methods normally need to employ huge number of the images (also including the corresponding reference images) to train their models,  and thus the confidence contribution of the reference color images encoded into the learned models in learning-based methods will be much lower than those exemplar-based methods. That will explain why the colorization results obtained by learning-based methods normally have some global color drifts in the whole image. In our hybrid approach, the knowledge for the matching operation and the transferring color space (the reference image) are decoupled comparing with those learning-based methods~\footnote{The learning-based methods can be viewed as coupling the knowledge of the matching operation and transferring color space into the same model.}, which means the training process in our method only need to learn (or encode) the matching relations under various structural composition of images, and for each individual target grayscale image that needs to be colorized, our method will also import strong prior knowledge from the corresponding reference images.

It is well known that when giving more training samples, learning-based method could have better generalization ability. As for He~\cite{he2018deep}, which also combines the learning-based method with the exemplar-based method, trained deep networks on approximately 0.7 million image pairs, which is 700 times as much as ours. While the colorization performance are comparable on most test images. Meanwhile, the quantitative experiment shows that our method has lower JND error. When giving the appropriate reference image beforehand, the runtime of our algorithm for colorizing a 256 $\times$ 256 image is about 0.2s, including 0.088s for feature extraction, 0.012s for feature matching and 0.1s for colorization. Which is slightly slower than 0.166s mentioned in~\cite{he2018deep} (including reference recommendation additionally). Considering that our algorithm does not using any GPU, which is more suitable for pure CPU application scenarios.

The main limitation of our approach may focus on the selection of the similar reference images. However, this problem may be beyond the scope of our works, as there are quite a lot of successfully image retrieval methods, such as~\cite{grauman2013learning}~\cite{tolias2016image}~\cite{do2017embedding}, which can select the optimal reference image for the input grayscale image automatically and precisely. The second limitaion may come from the super-pixels, which may cause color bleeding artifacts near the boundaries in the results.

\section{Conclusion}

In this paper, we propose an exemplar-based colorization method with a learning framework. Firstly, the sample images are clustered into several categories based on their GLCM based global features. And then we learn the optimal matching parameters for each category by minimizing the errors between the transferred image and the corresponding ground truths of all the images belonging to that category. Those learned parameters of each category are then used in the feature matching process to colorize the new coming target grayscale image. After transferring the colors from the reference image to the target image, a post-processing with spatial consistency is used to adjust the wrong matched colors. Finally, we obtain the colorized image through the color distribution process. Experimental results show that our method is effective and has comparable performance with the state-of-the-art methods both in visual sense and quantitative analysis.

The future work is going to further integrate the method to select the optimal reference color image for the target grayscale image automatically into our learning framework. With the help of GPU, we will also try to learn a larger number of sample images on a larger dataset and also implement a faster colorization in the next version.



\ifCLASSOPTIONcaptionsoff
  \newpage
\fi

\bibliographystyle{IEEEtran}
\bibliography{refs}

\begin{thebibliography}{10}
\providecommand{\url}[1]{#1}
\csname url@samestyle\endcsname
\providecommand{\newblock}{\relax}
\providecommand{\bibinfo}[2]{#2}
\providecommand{\BIBentrySTDinterwordspacing}{\spaceskip=0pt\relax}
\providecommand{\BIBentryALTinterwordstretchfactor}{4}
\providecommand{\BIBentryALTinterwordspacing}{\spaceskip=\fontdimen2\font plus
\BIBentryALTinterwordstretchfactor\fontdimen3\font minus
  \fontdimen4\font\relax}
\providecommand{\BIBforeignlanguage}[2]{{%
\expandafter\ifx\csname l@#1\endcsname\relax
\typeout{** WARNING: IEEEtran.bst: No hyphenation pattern has been}%
\typeout{** loaded for the language `#1'. Using the pattern for}%
\typeout{** the default language instead.}%
\else
\language=\csname l@#1\endcsname
\fi
#2}}
\providecommand{\BIBdecl}{\relax}
\BIBdecl

\bibitem{DBLP:journals/tsmc/LiGKH17}
\BIBentryALTinterwordspacing
Y.~Li, Y.~Guo, Y.~Kao, and R.~He, ``Image piece learning for weakly supervised
  semantic segmentation,'' \emph{{IEEE} Trans. Systems, Man, and Cybernetics:
  Systems}, vol.~47, no.~4, pp. 648--659, 2017. [Online]. Available:
  \url{https://doi.org/10.1109/TSMC.2016.2623683}
\BIBentrySTDinterwordspacing

\bibitem{DBLP:journals/tsmc/PengCYT17}
\BIBentryALTinterwordspacing
Q.~Peng, Y.~Cheung, X.~You, and Y.~Y. Tang, ``A hybrid of local and global
  saliencies for detecting image salient region and appearance,'' \emph{{IEEE}
  Trans. Systems, Man, and Cybernetics: Systems}, vol.~47, no.~1, pp. 86--97,
  2017. [Online]. Available: \url{https://doi.org/10.1109/TSMC.2016.2564922}
\BIBentrySTDinterwordspacing

\bibitem{yang2015graph}
J.~Yang, Z.~Gan, K.~Li, and C.~Hou, ``Graph-based segmentation for rgb-d data
  using 3-d geometry enhanced superpixels,'' \emph{IEEE transactions on
  cybernetics}, vol.~45, no.~5, pp. 927--940, 2015.

\bibitem{peng2016high}
J.~Peng, J.~Shen, and X.~Li, ``High-order energies for stereo segmentation,''
  \emph{IEEE transactions on cybernetics}, vol.~46, no.~7, pp. 1616--1627,
  2016.

\bibitem{sima2017bottom}
H.~Sima, P.~Guo, Y.~Zou, Z.~Wang, and M.~Xu, ``Bottom-up merging segmentation
  for color images with complex areas,'' \emph{IEEE Transactions on Systems,
  Man, and Cybernetics: Systems}, 2017.

\bibitem{DBLP:journals/tsmc/PassalisT17}
\BIBentryALTinterwordspacing
N.~Passalis and A.~Tefas, ``Learning neural bag-of-features for large-scale
  image retrieval,'' \emph{{IEEE} Trans. Systems, Man, and Cybernetics:
  Systems}, vol.~47, no.~10, pp. 2641--2652, 2017. [Online]. Available:
  \url{https://doi.org/10.1109/TSMC.2017.2680404}
\BIBentrySTDinterwordspacing

\bibitem{zhang2012automatic}
S.~Zhang, J.~Huang, H.~Li, and D.~N. Metaxas, ``Automatic image annotation and
  retrieval using group sparsity,'' \emph{IEEE Transactions on Systems, Man,
  and Cybernetics, Part B (Cybernetics)}, vol.~42, no.~3, pp. 838--849, 2012.

\bibitem{liu2016unsupervised}
L.~Liu, M.~Yu, and L.~Shao, ``Unsupervised local feature hashing for image
  similarity search,'' \emph{IEEE transactions on cybernetics}, vol.~46,
  no.~11, pp. 2548--2558, 2016.

\bibitem{zhang2017landmark}
X.~Zhang, S.~Wang, Z.~Li, and S.~Ma, ``Landmark image retrieval by jointing
  feature refinement and multimodal classifier learning,'' \emph{IEEE
  transactions on cybernetics}, 2017.

\bibitem{zong2015fast}
G.~Zong, Y.~Chen, G.~Cao, and J.~Dong, ``Fast image colorization based on local
  and global consistency,'' in \emph{Computational Intelligence and Design
  (ISCID), 2015 8th International Symposium on}, vol.~1.\hskip 1em plus 0.5em
  minus 0.4em\relax IEEE, 2015, pp. 366--369.

\bibitem{li2017example}
B.~Li, F.~Zhao, Z.~Su, X.~Liang, Y.-K. Lai, and P.~L. Rosin, ``Example-based
  image colorization using locality consistent sparse representation,''
  \emph{IEEE Transactions on Image Processing}, vol.~26, no.~11, pp.
  5188--5202, 2017.

\bibitem{Zhang:2017:RUI:3072959.3073703}
\BIBentryALTinterwordspacing
R.~Zhang, J.-Y. Zhu, P.~Isola, X.~Geng, A.~S. Lin, T.~Yu, and A.~A. Efros,
  ``Real-time user-guided image colorization with learned deep priors,''
  \emph{ACM Transaction on Graphics}, vol.~36, no.~4, pp. 119:1--119:11, Jul.
  2017. [Online]. Available: \url{http://doi.acm.org/10.1145/3072959.3073703}
\BIBentrySTDinterwordspacing

\bibitem{he2018deep}
M.~He, D.~Chen, J.~Liao, P.~V. Sander, and L.~Yuan, ``Deep exemplar-based
  colorization,'' \emph{ACM Transactions on Graphics (TOG)}, vol.~37, no.~4,
  p.~47, 2018.

\bibitem{Levin:2004}
\BIBentryALTinterwordspacing
A.~Levin, D.~Lischinski, and Y.~Weiss, ``Colorization using optimization,''
  \emph{ACM Trans. Graph.}, vol.~23, no.~3, pp. 689--694, Aug. 2004. [Online].
  Available: \url{http://doi.acm.org/10.1145/1015706.1015780}
\BIBentrySTDinterwordspacing

\bibitem{Yatziv:2004}
L.~Yatziv and G.~Sapiro, ``Fast image and video colorization using chrominance
  blending,'' \emph{IEEE TRANSACTIONS ON IMAGE PROCESSING}, vol.~15, p. 2006,
  2004.

\bibitem{Huang:2005}
\BIBentryALTinterwordspacing
Y.-C. Huang, Y.-S. Tung, J.-C. Chen, S.-W. Wang, and J.-L. Wu, ``An adaptive
  edge detection based colorization algorithm and its applications,'' in
  \emph{Proceedings of the 13th Annual ACM International Conference on
  Multimedia}, ser. MULTIMEDIA '05.\hskip 1em plus 0.5em minus 0.4em\relax New
  York, NY, USA: ACM, 2005, pp. 351--354. [Online]. Available:
  \url{http://doi.acm.org/10.1145/1101149.1101223}
\BIBentrySTDinterwordspacing

\bibitem{Luan:2007}
\BIBentryALTinterwordspacing
Q.~Luan, F.~Wen, D.~Cohen-Or, L.~Liang, Y.-Q. Xu, and H.-Y. Shum, ``Natural
  image colorization,'' in \emph{Proceedings of the 18th Eurographics
  Conference on Rendering Techniques}, ser. EGSR'07.\hskip 1em plus 0.5em minus
  0.4em\relax Aire-la-Ville, Switzerland, Switzerland: Eurographics
  Association, 2007, pp. 309--320. [Online]. Available:
  \url{http://dx.doi.org/10.2312/EGWR/EGSR07/309-320}
\BIBentrySTDinterwordspacing

\bibitem{Welsh:2002}
\BIBentryALTinterwordspacing
T.~Welsh, M.~Ashikhmin, and K.~Mueller, ``Transferring color to greyscale
  images,'' \emph{ACM Trans. Graph.}, vol.~21, no.~3, pp. 277--280, Jul. 2002.
  [Online]. Available: \url{http://doi.acm.org/10.1145/566654.566576}
\BIBentrySTDinterwordspacing

\bibitem{Irony:2005}
\BIBentryALTinterwordspacing
R.~Irony, D.~Cohen-Or, and D.~Lischinski, ``Colorization by example,'' in
  \emph{Proceedings of the Sixteenth Eurographics Conference on Rendering
  Techniques}, ser. EGSR '05.\hskip 1em plus 0.5em minus 0.4em\relax
  Aire-la-Ville, Switzerland, Switzerland: Eurographics Association, 2005, pp.
  201--210. [Online]. Available:
  \url{http://dx.doi.org/10.2312/EGWR/EGSR05/201-210}
\BIBentrySTDinterwordspacing

\bibitem{Charpiat:2008}
\BIBentryALTinterwordspacing
G.~Charpiat, M.~Hofmann, and B.~Sch\"{o}lkopf, ``Automatic image colorization
  via multimodal predictions,'' in \emph{Proceedings of the 10th European
  Conference on Computer Vision: Part III}, ser. ECCV '08.\hskip 1em plus 0.5em
  minus 0.4em\relax Berlin, Heidelberg: Springer-Verlag, 2008, pp. 126--139.
  [Online]. Available: \url{http://dx.doi.org/10.1007/978-3-540-88690-7\_10}
\BIBentrySTDinterwordspacing

\bibitem{Liu:2008}
\BIBentryALTinterwordspacing
X.-P. Liu, L.~Wan, Y.-G. Qu, T.-T. Wong, S.~Lin, C.-S. Leung, and P.-A. Heng,
  ``Intrinsic colorization,'' \emph{ACM Trans. Graph.}, vol.~27, no.~5, pp.
  152:1--152:9, Dec. 2008. [Online]. Available:
  \url{http://doi.acm.org/10.1145/1409060.1409105}
\BIBentrySTDinterwordspacing

\bibitem{Gupta:2012}
\BIBentryALTinterwordspacing
K.~R. Gupta, A.~Y.-S. Chia, D.~Rajan, E.-S. Ng, and Z.-Y. Huang, ``Image
  colorization using similar images,'' in \emph{Proceedings of the 20th ACM
  International Conference on Multimedia}, ser. MM '12.\hskip 1em plus 0.5em
  minus 0.4em\relax New York, NY, USA: ACM, 2012, pp. 369--378. [Online].
  Available: \url{http://doi.acm.org/10.1145/2393347.2393402}
\BIBentrySTDinterwordspacing

\bibitem{pierre2015luminance}
F.~Pierre, J.-F. Aujol, A.~Bugeau, N.~Papadakis, and V.-T. Ta,
  ``Luminance-chrominance model for image colorization,'' \emph{SIAM Journal on
  Imaging Sciences}, vol.~8, no.~1, pp. 536--563, 2015.

\bibitem{deshpande2015learning}
A.~Deshpande, J.~Rock, and D.~Forsyth, ``Learning large-scale automatic image
  colorization,'' in \emph{Proceedings of the IEEE International Conference on
  Computer Vision}, 2015, pp. 567--575.

\bibitem{varga2016fully}
D.~Varga and T.~Szir{\'a}nyi, ``Fully automatic image colorization based on
  convolutional neural network,'' in \emph{Pattern Recognition (ICPR), 2016
  23rd International Conference on}.\hskip 1em plus 0.5em minus 0.4em\relax
  IEEE, 2016, pp. 3691--3696.

\bibitem{larsson2016learning}
G.~Larsson, M.~Maire, and G.~Shakhnarovich, ``Learning representations for
  automatic colorization,'' in \emph{European Conference on Computer
  Vision}.\hskip 1em plus 0.5em minus 0.4em\relax Springer, 2016, pp. 577--593.

\bibitem{zhang2016colorful}
R.~Zhang, P.~Isola, and A.~A. Efros, ``Colorful image colorization,'' in
  \emph{European Conference on Computer Vision}.\hskip 1em plus 0.5em minus
  0.4em\relax Springer, 2016, pp. 649--666.

\bibitem{iizuka2016let}
S.~Iizuka, E.~Simo-Serra, and H.~Ishikawa, ``Let there be color!: joint
  end-to-end learning of global and local image priors for automatic image
  colorization with simultaneous classification,'' \emph{ACM Transactions on
  Graphics (TOG)}, vol.~35, no.~4, p. 110, 2016.

\bibitem{cheng2017colorization}
Z.~Cheng, Q.~Yang, and B.~Sheng, ``Colorization using neural network
  ensemble,'' \emph{IEEE Transactions on Image Processing}, vol.~26, no.~11,
  pp. 5491--5505, 2017.

\bibitem{zhang2017real}
R.~Zhang, J.-Y. Zhu, P.~Isola, X.~Geng, A.~S. Lin, T.~Yu, and A.~A. Efros,
  ``Real-time user-guided image colorization with learned deep priors,''
  \emph{arXiv preprint arXiv:1705.02999}, 2017.

\bibitem{xiao2019interactive}
Y.~Xiao, P.~Zhou, Y.~Zheng, and C.-S. Leung, ``Interactive deep colorization
  using simultaneous global and local inputs,'' in \emph{ICASSP 2019-2019 IEEE
  International Conference on Acoustics, Speech and Signal Processing
  (ICASSP)}.\hskip 1em plus 0.5em minus 0.4em\relax IEEE, 2019, pp. 1887--1891.

\bibitem{han2016two}
J.~Han, D.~Zhang, S.~Wen, L.~Guo, T.~Liu, and X.~Li, ``Two-stage learning to
  predict human eye fixations via sdaes,'' \emph{IEEE transactions on
  cybernetics}, vol.~46, no.~2, pp. 487--498, 2016.

\bibitem{Bugeau:2014}
A.~Bugeau, V.-T. Ta, and N.~Papadakis, ``Variational exemplar-based image
  colorization,'' \emph{IEEE Transactions on Image Processing}, vol.~23, pp.
  298--307, 2014.

\bibitem{cheng2015deep}
Z.~Cheng, Q.~Yang, and B.~Sheng, ``Deep colorization,'' in \emph{Proceedings of
  the IEEE International Conference on Computer Vision}, 2015, pp. 415--423.

\bibitem{russakovsky2015imagenet}
O.~Russakovsky, J.~Deng, H.~Su, J.~Krause, S.~Satheesh, S.~Ma, Z.~Huang,
  A.~Karpathy, A.~Khosla, M.~Bernstein \emph{et~al.}, ``Imagenet large scale
  visual recognition challenge,'' \emph{International Journal of Computer
  Vision}, vol. 115, no.~3, pp. 211--252, 2015.

\bibitem{xiaoexample}
C.~Xiao, C.~Han, Z.~Zhang, J.~Qin, T.-T. Wong, G.~Han, and S.~He,
  ``Example-based colourization via dense encoding pyramids,'' in
  \emph{Computer Graphics Forum}.\hskip 1em plus 0.5em minus 0.4em\relax Wiley
  Online Library.

\bibitem{Haralick:1973}
R.~M. Haralick, K.~Shanmugam, and I.~Dinstein, ``{Textural Features for Image
  Classification},'' \emph{IEEE Transactions on Systems, Man, and Cybernetics},
  vol.~3, pp. 610--621, 1973.

\bibitem{Xiao:2007}
J.-X. Xiao, J.-D. Wang, P.~Tan, and L.~Quan, ``Joint affinity propagation for
  multiple view segmentation,'' in \emph{In ICCV}, 2007, p.~43.

\bibitem{Levinshtein:2009}
\BIBentryALTinterwordspacing
A.~Levinshtein, A.~Stere, K.~N. Kutulakos, D.~J.Fleet, S.~J. Dickinson, and
  K.~Siddiqi, ``Turbopixels: Fast superpixels using geometric flows,''
  \emph{IEEE Trans. Pattern Anal. Mach. Intell.}, vol.~31, no.~12, pp.
  2290--2297, Dec. 2009. [Online]. Available:
  \url{http://dx.doi.org/10.1109/TPAMI.2009.96}
\BIBentrySTDinterwordspacing

\bibitem{Manjunath:1996}
\BIBentryALTinterwordspacing
B.~S. Manjunath and W.~Y. Ma, ``Texture features for browsing and retrieval of
  image data,'' \emph{IEEE Trans. Pattern Anal. Mach. Intell.}, vol.~18, no.~8,
  pp. 837--842, Aug. 1996. [Online]. Available:
  \url{http://dx.doi.org/10.1109/34.531803}
\BIBentrySTDinterwordspacing

\bibitem{Bay:2008}
\BIBentryALTinterwordspacing
H.~Bay, A.~Ess, T.~Tuytelaars, and L.~V. Gool, ``Speeded-up robust features
  (surf),'' \emph{Computer Vision and Image Understanding}, vol. 110, no.~3,
  pp. 346 -- 359, 2008, similarity Matching in Computer Vision and Multimedia.
  [Online]. Available:
  \url{http://www.sciencedirect.com/science/article/pii/S1077314207001555}
\BIBentrySTDinterwordspacing

\bibitem{Manolis:2005}
L.~Manolis, ``A brief description of the levenberg-marquardt algorithm
  implemented by levmar,'' \emph{Foundation of Research and Technology},
  vol.~4, pp. 1--6, 2005.

\bibitem{Lafferty01conditionalrandom}
J.~Lafferty, ``Conditional random fields: Probabilistic models for segmenting
  and labeling sequence data.''\hskip 1em plus 0.5em minus 0.4em\relax Morgan
  Kaufmann, 2001, pp. 282--289.

\bibitem{krahenbuhl2015data}
P.~Kr{\"a}henb{\"u}hl, C.~Doersch, J.~Donahue, and T.~Darrell, ``Data-dependent
  initializations of convolutional neural networks,'' \emph{arXiv preprint
  arXiv:1511.06856}, 2015.

\bibitem{sharma2005ciede2000}
G.~Sharma, W.~Wu, and E.~N. Dalal, ``The ciede2000 color-difference formula:
  Implementation notes, supplementary test data, and mathematical
  observations,'' \emph{Color Research \& Application}, vol.~30, no.~1, pp.
  21--30, 2005.

\bibitem{bala2003digital}
R.~Bala and G.~Sharma, ``Digital color imaging handbook,'' 2003.

\bibitem{mahy1994evaluation}
M.~Mahy, L.~Van~Eycken, and A.~Oosterlinck, ``Evaluation of uniform color
  spaces developed after the adoption of cielab and cieluv,'' \emph{rn},
  vol.~2, p. 1nrW, 1994.

\bibitem{grauman2013learning}
K.~Grauman and R.~Fergus, ``Learning binary hash codes for large-scale image
  search,'' in \emph{Machine learning for computer vision}.\hskip 1em plus
  0.5em minus 0.4em\relax Springer, 2013, pp. 49--87.

\bibitem{tolias2016image}
G.~Tolias, Y.~Avrithis, and H.~J{\'e}gou, ``Image search with selective match
  kernels: aggregation across single and multiple images,'' \emph{International
  Journal of Computer Vision}, vol. 116, no.~3, pp. 247--261, 2016.

\bibitem{do2017embedding}
T.-T. Do and N.-M. Cheung, ``Embedding based on function approximation for
  large scale image search,'' \emph{IEEE Transactions on Pattern Analysis and
  Machine Intelligence}, 2017.

\end{thebibliography}


\end{document}